\documentclass{article} 
\usepackage{iclr2025_conference,times}

\usepackage{amsmath,amsfonts,bm}

\def\eqref#1{equation~\ref{#1}}

\def\1{\bm{1}}

\DeclareMathAlphabet{\mathsfit}{\encodingdefault}{\sfdefault}{m}{sl}
\SetMathAlphabet{\mathsfit}{bold}{\encodingdefault}{\sfdefault}{bx}{n}

\usepackage{hyperref}
\usepackage{url}

\usepackage{makecell}
\usepackage{array}
\usepackage{graphicx} 
\usepackage{booktabs}
\usepackage{multirow} 
\usepackage{colortbl}
\usepackage{xcolor}   
\usepackage{threeparttable}
\usepackage{wrapfig}
\usepackage{subcaption}
\usepackage{algorithm}
\usepackage{algpseudocode}
\usepackage{listings}
\usepackage{amsthm}

\usepackage{bbm}
\usepackage{amssymb}
\usepackage{enumitem}
\usepackage{pifont}
\usepackage{tikz}
\usetikzlibrary{shadows}       
\usetikzlibrary{shadows.blur}
\usetikzlibrary{arrows.meta, positioning, shapes, calc, fit}
\usepackage{tabularx}
\newcommand{\name}{Muddit}

\definecolor{lightBlue}{RGB}{173, 216, 230}
\definecolor{lightGreen}{RGB}{204, 255, 204}

\definecolor{Blue}{RGB}{0, 183, 133}
\definecolor{Aquamarine}{RGB}{127, 255, 212}
\definecolor{Sepia}{RGB}{112, 66, 20}
\definecolor{BrickRed}{RGB}{203, 65, 84}
\colorlet{my-red}{BrickRed!90!Sepia}
\colorlet{my-blue}{Aquamarine!30!Blue}
\hypersetup{
  colorlinks,
  urlcolor  = blue,
  linkcolor = blue,
  citecolor = blue,
}
\newcommand{\blfootnote}[1]{\begingroup
\renewcommand\thefootnote{}\footnote{#1}\addtocounter{footnote}{-1}
\endgroup}
\usepackage{float}
\usepackage{marvosym}

\title{Muddit: Liberating Generation Beyond Text-to-Image with a Unified Discrete Diffusion Model}

\author{Qingyu Shi$^{*}$, Jinbin Bai$^{*\dagger}$, Zhuoran Zhao, Wenhao Chai, Kaidong Yu, \\
\textbf{Jianzong Wu, Yunhai Tong, Xiangtai Li$^{\ddagger}$, Xuelong Li,  Shuicheng Yan$^{\ddagger}$} \vspace{5pt} \\ 
{$^*$Equal Contribution,} 
{$^\dagger$Project Lead,}
{$^\ddagger$Corresponding Authors} \vspace{5pt} \\
{\centering \texttt{Model: \url{https://huggingface.co/MeissonFlow/Muddit}}}
  \\
   {\centering \texttt{Code: ~\url{https://github.com/M-E-AGI-Lab/Muddit}}}
}

\iclrfinalcopy 
\begin{document}
\maketitle

\blfootnote{$^{1}$Peking University, $^{2}$National University of Singapore, $^{3}$Princeton University}
\blfootnote{\Letter:~ \texttt{jinbin.bai@u.nus.edu}}

\begin{abstract}

Unified generation models aim to handle diverse tasks across modalities—such as text generation, image generation, and vision-language reasoning—within a single architecture and decoding paradigm. 
Autoregressive unified models suffer from slow inference due to sequential decoding, and non-autoregressive unified models suffer from weak generalization due to limited pretrained backbones. 
We introduce the second-generation Meissonic: \name{}, a \textbf{u}nified \textbf{d}iscrete \textbf{di}ffusion \textbf{t}ransformer that enables fast and parallel generation across both text and image modalities. 
Unlike prior unified diffusion models trained from scratch, \name{} integrates strong \textbf{visual priors} from a pretrained text-to-image backbone with a lightweight text decoder, enabling flexible and high-quality multimodal generation under a unified architecture.
Empirical results show that \name{} achieves competitive or superior performance compared to significantly larger autoregressive models in both quality and efficiency.
The work highlights the potential of purely discrete diffusion, when equipped with strong visual priors, as a scalable and effective backbone for unified generation.

\end{abstract}
\section{Introduction}
\label{sec:intro}

\begin{figure}[t]
    \centering
    \includegraphics[width=1.\linewidth]{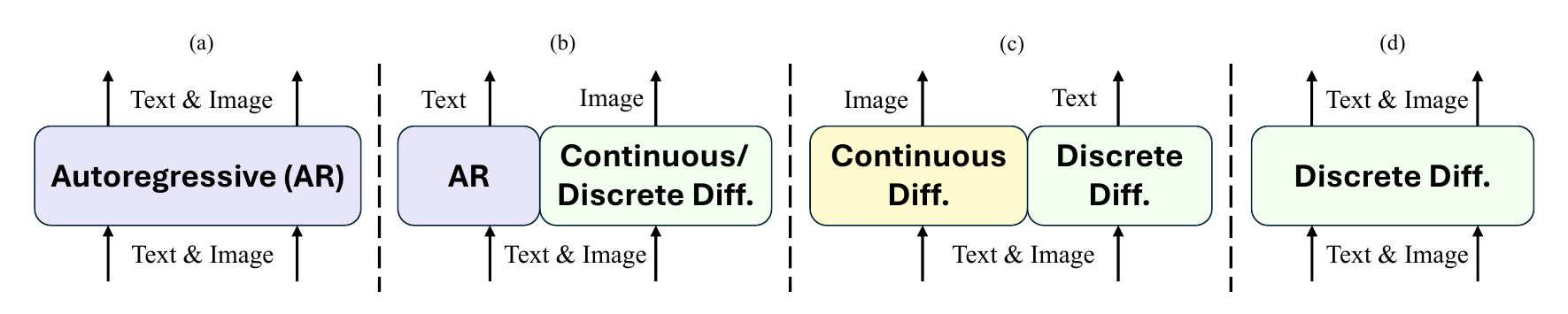}
    \caption{Four types of unified generative models. More details can be found in Sec.~\ref{sec:related_work}.}
    \label{fig:unified_cls}
    \vspace{-10pt}
\end{figure}

Multimodal generative models capable of handling both text and images have rapidly advanced, typically relying on large autoregressive (AR) Transformers, also known as large language models (LLMs)~\citep{touvron2023llama}.
These unified models represent both modalities as token sequences and generate outputs in a left-to-right autoregressive manner.
However, this sequential decoding imposes a major inference bottleneck.
For instance, in early unified transformers~\citep{sun2024autoregressive}, as illustrated in Fig.~\ref{fig:unified_cls}(a), generating a single image requires sampling thousands of visual tokens one at a time. 
Despite strong correlation among adjacent image tokens, each token prediction triggers a full network forward, resulting in significant redundant computation.
As a result, inference becomes \textbf{extremely slow and compute-intensive}.
We refer to this as the first ``dark cloud” over current unified generative models.
Moreover, AR decoding enforces a rigid generation order. This prevents speed-quality trade-offs or flexible conditional generation like inpainting without fine-tuning, which severely limits practical applicability in interactive or real-time scenarios. 
To mitigate these limitations, some hybrid approaches~\citep{chen2025blip3ofamilyfullyopen,pan2025transfer,chen2025multimodal}, adopt AR language models paired with diffusion-based image synthesis heads (Fig.~\ref{fig:unified_cls}(b)).
However, these “glue” architectures fall short of true unification, as they lack a shared generative modeling paradigm across modalities.

Recent work like Dual-Diffusion~\citep{li2024dual} and Diffuse Everything~\citep{rojas2025diffuse} (Fig.~\ref{fig:unified_cls}(c)) claims to unify modalities under discrete diffusion, but they ultimately rely on continuous diffusion for image generation. This fundamental mismatch in generative principles undermines their claim of true unification. 
UniDisc~\citep{swerdlow2025unified}~(Fig.~\ref{fig:unified_cls}(d)), takes a more promising step by applying discrete diffusion\footnote{MaskGIT, MaskAR, RandomAR, and Discrete Diffusion share significant conceptual and practical overlaps, often differing only in decoding order or architectural nuances. We elaborate on their connections in the next section.
While Meissonic~\citep{bai2024meissonic} follows the naming convention of MaskGIT~\citep{chang2022maskgit}, we standardize terminology in this paper by referring to all such models under the umbrella of Discrete Diffusion.}  over unified token spaces. This allows parallel refinement of text and image tokens, improving inference efficiency and enabling more flexible conditioning. 
However, the overall generation quality of UniDisc remains far from satisfactory. 
For example, it struggles to produce high-resolution $1024 \times 1024$  images, fails to match the fidelity of early diffusion models such as Stable Diffusion 1.5, and lacks support for vision-language reasoning tasks such as visual question answering (VQA). These limitations expose the second “dark cloud”: \textbf{the absence of strong pre-trained discrete diffusion backbone models}: 
Unlike established unified autoregressive models that leverage powerful pretrained large language models, current unified discrete diffusion models are typically trained from scratch on mixed-modality tokens, which limits both their generative fidelity and transferability. 
Without modular components carrying rich pixel-level priors, these models face generalization and scalability bottlenecks.

Taken together, the two dark clouds: inefficient autoregressive sampling and the lack of strong pretrained foundations, highlight the need for a new generation of unified models. In this work, we present the second-generation Meissonic: \textbf{\name{}}, a \textbf{M}askGIT-style \textbf{u}nified \textbf{d}iscrete \textbf{di}ffusion \textbf{t}ransformer equipped with a lightweight text decoder. By combining the strengths of parallel discrete diffusion and semantically rich image priors from a pre-trained Meissonic text-to-image backbone~\citep{bai2024meissonic}, \name{} enables scalable, efficient, and flexible sampling while significantly improving alignment and quality across modalities and various tasks such as high-resolution text-to-image synthesis, image-to-text synthesis, and visual question answering. 
We systematically detail the training objective of unified discrete diffusion models, the masking strategy, and the shared inference sampling strategy across three tasks. Finally, we conduct comprehensive evaluations with current popular unified models on several benchmarks, including GenEval, CIDEr, VQAv2, MME, and GQA, demonstrating \name{}'s superior performance and efficiency, validating that the unexplored purely discrete diffusion approach can rival, or even surpass, much larger autoregressive-based unified models. 
While concurrent unified generation models~\citep{yang2025mmada} often build upon a language modeling prior—leveraging pretrained dLLMs as the backbone—we instead take a visual-first approach. Muddit is built upon an image generation prior, offering a new path toward unifying vision and language tasks within a discrete diffusion framework. We hope that this work inspires a new trend for unified generative modeling, grounded in discrete diffusion, beyond the boundaries of traditional text-to-image synthesis~\citep{bai2024meissonic} and text synthesis~\citep{nie2025large,InceptionLabs2025}.

\section{Method}
\label{sec:method}

\subsection{Discrete Diffusion with Unified Image and Text Perspective}

In discrete diffusion, a sample $x \in \mathcal{X}$ is treated as a one-hot vector $\mathbf{x}$, where $\mathcal{X}=\{1,\dots, N\}$.  
For language models, $N$ equals the vocabulary size. While for image models, $N$ is the number of discrete image token IDs obtained from a tokenizer or VQ codebook.
At each diffusion step, we stochastically corrupt the tokens, gradually transforming the data distribution into a maximally entropic categorical prior; the generative model then learns to invert this corruption. 
Following recent works~\citep{SEDD, bai2024meissonic} that cast token corruption as a continuous–time Markov chain (CTMC) over the finite alphabet $\mathcal{X}$, we let
\begin{equation}
\frac{d\,p_t}{dt}=Q_t\,p_t ,
\label{eq:ctmc_forward}
\end{equation}
where $p_t \in \mathbb{R}^{N+1}$ is the distribution of $x_t$, and the time–dependent matrix $Q_t$ transports the data distribution $p_{0}\approx p_{\mathrm{data}}$ to the maximally entropic “noise’’ distribution $p_{1} = p_{\mathrm{stationary}}$. 
We adopt the absorbing-state (masked) diffusion variant that has proved particularly effective in text modelling: every symbol can jump to a dedicated mask token ${\tt m} = (\underbrace{0, \dots,0}_{N}, 1)$ but never leaves it, i.e.\ ${\tt m}$ is an absorbing class.

\noindent
\textbf{Forward posterior.}
Marginalizing $\mathbf{x}$ gives
\begin{equation}
q(x_t\mid\mathbf{x})
= \operatorname{Cat}\!\bigl(x_t\mid\alpha_t\mathbf{x}+(1-\alpha_t){\tt m}\bigr).
\label{eq:ctmc_posterior}
\end{equation}
$\operatorname{Cat}(\cdot)$ denotes a categorical distribution; it returns a one-hot token sampled from the probability vector inside the parentheses.  
$\alpha_t \in [0, 1]$ is the \emph{survival probability}, \textit{i.e.}\ the probability that an individual token has not yet been masked by time~$t$. Thus $x_t$ equals the original clean token with probability $\alpha_t$ and equals the mask token ${\tt m}$ with probability $1-\alpha_t$.

\noindent
\textbf{Reverse process.}
For any $0<s<t<1$, the CTMC induces an analytic posterior
\begin{equation}
q(x_s\mid x_t,\mathbf{x})=
\begin{cases}
\operatorname{Cat}(x_s\mid x_t), & x_t\neq{\tt m},\\[6pt]
\operatorname{Cat}\!\Bigl(x_s\mid \dfrac{(1-\alpha_s){\tt m}+(\alpha_s-\alpha_t)\mathbf{x}}
                                   {1-\alpha_t}\Bigr), & x_t={\tt m},
\end{cases}
\label{eq:ctmc_reverse}
\end{equation}
$x_t$ and $x_s$ are the corrupted tokens at times $t$ and $s$ ($s<t$).  
If $x_t$ is already a real vocabulary token ($x_t\neq{\tt m}$) it stays unchanged going backwards; otherwise, when $x_t={\tt m}$, the distribution over $x_s$ is a convex combination of the mask and the clean token $\mathbf{x}$, weighted by their respective survival probabilities $\alpha_s$ and $\alpha_t$.

\noindent
\textbf{Training Objective.}
We employ a masked‐token predictor $x_\theta(x_t,\alpha_t)\!\approx\!\mathbf{x}$, which leads to the continuous‐time negative ELBO~\citep{ou2024your,sahoo2024simple,md4}
\begin{equation}
\mathcal{L}_{\mathrm{NELBO}}
= \mathbb{E}_{q(x_t\mid\mathbf{x})}\;
  \Bigl[
    \int_{0}^{1}
    \frac{\alpha_t'}{1-\alpha_t}\,
    \log\bigl(x_\theta(x_t,\alpha_t)\!\cdot\!\mathbf{x}\bigr)\,dt
  \Bigr],
\label{eq:nelbo_ctmc}
\end{equation}
where $\alpha_t'=\tfrac{d\alpha_t}{dt}$ and $\mathbf{x}$ is the one-hot vector of ground truth.
$x_\theta(x_t,\alpha_t)\!\in\!\mathbb{R}^{N+1}$ is the model’s predicted categorical probability vector for the clean token given the corrupted input $(x_t,\alpha_t)$. 

During generation, we start from an all‐mask sequence ($t=1$) and integrate the reverse CTMC towards $t=0$, repeatedly replacing every masked position with the model’s categorical prediction. Because the corruption schedule and objective are \emph{identical} for any discrete alphabet $\mathcal{X}$, the same diffusion backbone unifies text and image generation. In the following section, we present \name{}, a unified framework that leverages discrete diffusion to model the generation tasks for both text and image jointly.

\subsection{Muddit}

\subsubsection{Unified Architecture}

\begin{figure}[t]
    \centering
    \includegraphics[width=1.\linewidth]{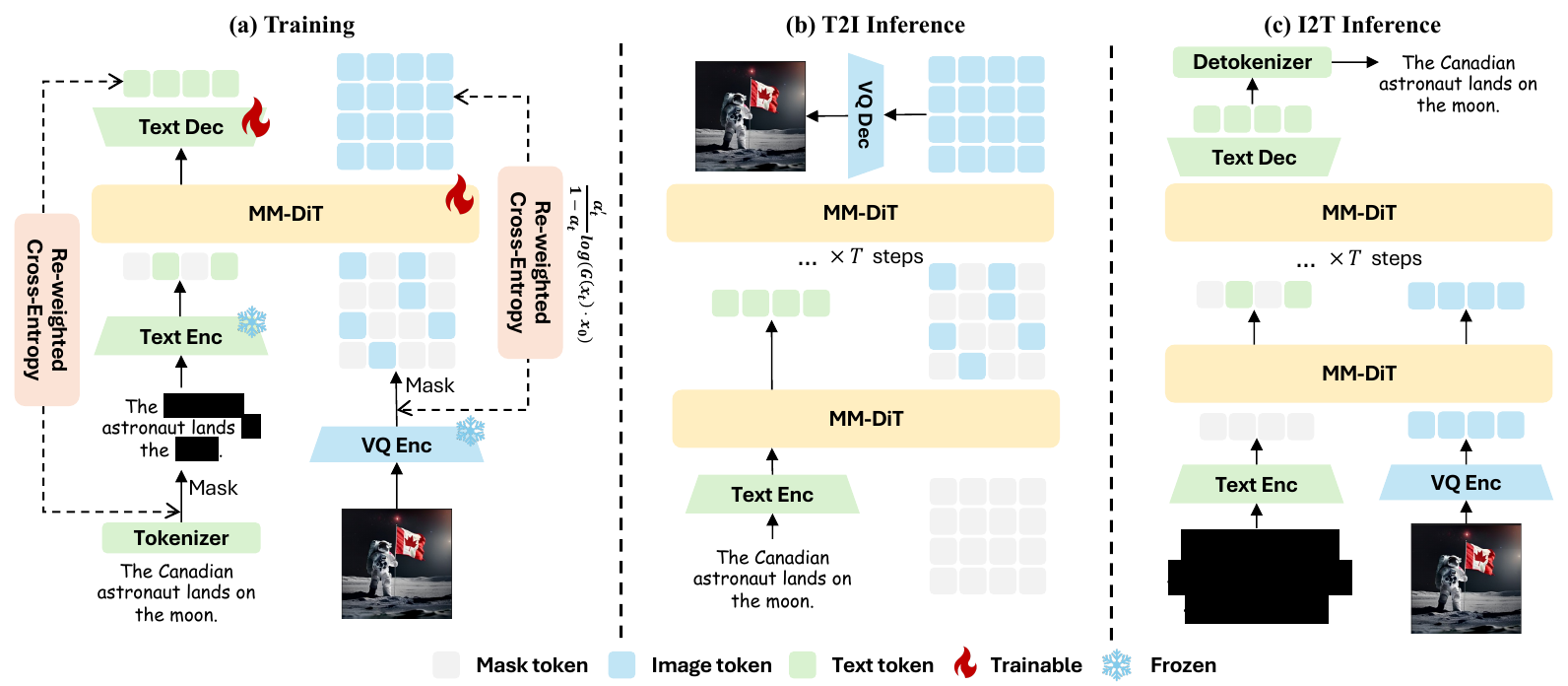}
    \caption{The training and inference architecture of \name{}. \textbf{(a)} During training, we randomly mask tokens from one of the two modalities. MM-DiT is trained to predict the masked tokens using a re-weighted cross-entropy loss, which jointly optimizes both the MM-DiT backbone and a lightweight text decoder. \textbf{(b)} In text-to-image inference, we initialize the image latent features using all-masked tokens and iteratively predict each latent token via MM-DiT. \textbf{(c)} In image-to-text inference, we similarly initialize all text tokens as masked and generate the text through the same iterative decoding process. Specifically for VQA tasks, we append mask token IDs to the end of the question and predict all masked token IDs as the final answer.}
    \label{fig:arch}
    \vspace{-10pt}
\end{figure}

As shown in Fig.~\ref{fig:arch}, our architecture comprises a text encoder $\mathtt{E_{txt}}$, image encoder \( \mathtt{E_{img}} \), transformer generator \( \mathtt{G} \), sampler \( \mathtt{S} \), text decoder \( \mathtt{D_{txt}} \), and image decoder \( \mathtt{D_{img}} \).
The generator \( \mathtt{G} \) is a single MM-DiT model, following the dual-/single-stream design of FLUX~\citep{Flux}.
Importantly, the generator $\mathtt{G}$ is initialized from Meissonic~\citep{bai2024meissonic}, which has been extensively trained for high-resolution text-to-image generation. This initialization brings in a strong pretrained image prior, capturing rich spatial structures and semantic correlations across image and text tokens, which significantly enhances sample quality and accelerates convergence in the multimodal setting.
Consequently, the same MM-DiT predicts the masked tokens for both modalities, which produces a shared generator for text and image synthesis.

To reduce the computational cost of high-resolution imagery and lengthy captions, we quantize both modalities into a compact discrete space. 
A pre-trained VQ-VAE acts as the image encoder $\mathtt{E_{\text{img}}}$, mapping pixels to codebook indices, while the CLIP text model, as $\mathtt{E_{\text{txt}}}$, provides the text token embeddings. 
The MM-DiT predicts clean tokens in this shared space, which a lightweight linear head $\mathtt{D_{\text{txt}}}$ converts back to text tokens.

\subsubsection{Unified Training}
\noindent
\textbf{Masking strategy.}
We model the forward posterior in Eq.~\ref{eq:ctmc_posterior} of both modalities using time-dependent hyperparameters $\alpha_t$, with the mask ratio defined as $\gamma_t = 1 - \alpha_t$. While BERT~\citep{devlin2018bert} employs a fixed mask ratio of 15\%, this setting is suitable for token completion but insufficient for generation. To support generative tasks, the design of $\gamma_t$ must satisfy the following criteria:
\begin{enumerate}
    \item $\gamma_t$ must be a continuous function, bounded between $0$ and $1$, for $t \in [0, 1]$. 
    \item $\gamma_t$ should monotonically decrease with respect to $t$, with boundary conditions $\gamma_0 \to 0$ (initially clean data) and $\gamma_1 \to 1$ (masking all tokens).  
\end{enumerate}
Several strategies for masking and sampling have been proposed to meet these criteria~\citep{chang2022maskgit}. We adopt a \textit{cosine scheduling strategy}.
During training, a timestep $t \in [0, 1]$ is sampled from a truncated $\arccos$ distribution, with the density function:  
\begin{equation}
    \gamma_t = \frac{2}{\pi} (1 - (1-t)^2)^{-\frac{1}{2}}.
\label{eq:gamma}
\end{equation}

During training, a mask ratio $\gamma_t \in [0, 1)$ is randomly sampled for each modality $\mathtt{x_0}$ (either image or text tokens), and the forward process (Eq.~\ref{eq:ctmc_posterior}) is applied by randomly replacing clean tokens with mask tokens to obtain $x_t$.

\noindent
\textbf{Unified training objective.}
Let $\mathbf{c}$ denote the conditioning: the text embedding when synthesizing an image, or the image embedding when generating a caption.
We randomly sample a mask ratio by Eq.~\ref{eq:gamma}. Then we corrupt the target sequence $\mathbf{x_0}$ (image or text tokens) with the CTMC described in Eq.~\ref{eq:ctmc_forward} and train a single masked-token predictor $\mathtt{G}(x_t,\alpha_t,\mathbf{c})$ to reconstruct $\mathbf{x_0}$. Both directions—text $\!\to$ image and image $\!\to$ text—share the identical continuous-time negative ELBO
\begin{equation}
\mathcal{L}_{\mathrm{unified}}
  = \mathbb{E}_{q(x_t \mid \mathbf{x})}\;
    \Bigl[
      \int_{0}^{1}
        \frac{\alpha_t'}{1-\alpha_t}\,
        \log\!\bigl( \mathtt{G}(x_t,\alpha_t,\mathbf{c}) \!\cdot\! \mathbf{x} \bigr)\,dt
    \Bigr],
\label{eq:unified_nelbo}
\end{equation}
where all symbols are as in Eq.~\ref{eq:nelbo_ctmc} but the $\mathtt{G}$ now receives the cross-modal condition~$\mathbf{c}$ as an additional input. \textbf{Key point:} switching from text $\!\to$ image to image $\!\to$ text merely changes the conditioning signal $\mathbf{c}$; the loss Eq.~\ref{eq:unified_nelbo} itself is unchanged. This symmetry keeps optimization identical across tasks and allows us to train a single parameter set jointly for both generation directions. During inference we again start from an all-mask sequence ($t{=}1$) and integrate the reverse CTMC towards $t{=}0$, feeding in the desired condition~$\mathbf{c}$ to obtain either an image or a sentence from the same diffusion backbone.

\subsubsection{Unified inference}
\noindent
\textbf{Sampling strategy.}
During inference, we apply the time-reversed posterior as defined in Eq.~\ref{eq:ctmc_reverse}.
\begin{equation}
\mathtt{S}(\mathtt{G}, x_t, t) =  p_{\theta}(x_s\mid x_t)=
\begin{cases}
\operatorname{Cat}(x_s\mid x_t), & x_t\neq{\tt m},\\[6pt]
\operatorname{Cat}\!\Bigl(x_s\mid \dfrac{(1-\alpha_s){\tt m}+(\alpha_s-\alpha_t)\mathtt{G}(x_t,\alpha_t, \mathtt{c})}
                                   {1-\alpha_t}\Bigr), & x_t={\tt m},
\end{cases}
\label{eq:unified_reverse}
\end{equation}

where $\theta$ denotes the parameters of $\mathtt{G}$, $\mathtt{c}$ is the multimodal condition, and $\alpha_t$ in Eq.~\ref{eq:gamma} is applied sequentially with $t$ taking values $1, \frac{T-1}{T}, \dots, \frac{1}{T}$, where $T$ is the total number of reverse steps. At each timestep $t$, Muddit predicts a fraction $\gamma_{t+\frac{1}{T}} - \gamma_{t}$ of the masked tokens by $\mathtt{G}$ and updates the masked tokens $\mathtt{x_t}$ by $\mathtt{S}$ , continuing iteratively until all masked tokens are recovered. 
This dynamic approach offers several advantages over autoregressive methods, which require the model to learn conditional probabilities $P(x_i \mid x_{<i})$ based on a fixed token ordering. In contrast, random masking with a variable ratio enables the model to learn $P(x_i \mid x_\Lambda)$, where $\Lambda$ denotes an arbitrary subset of observed tokens. This flexibility is essential for parallel sampling, allowing multiple tokens to be predicted simultaneously rather than sequentially. 

Our \name{} supports three tasks with a single generator \( \mathtt{G} \) and sampler \( \mathtt{S} \):
(i) text $\!\to$ image,
(ii) image $\!\to$ text (captioning),
and (iii) visual–question answering (VQA).
The only change across tasks is the conditioning source $\mathbf{c}$ provided to \( \mathtt{G} \); the diffusion process and guidance logic are shared.

\noindent
\textbf{(i) Text\,$\boldsymbol{\to}$\,image.}  
Given a text prompt \( \mathtt{tp} \!\in\!\mathcal{T} \),
the text encoder \( \mathtt{E_{txt}} \) produces a text token embedding \(\mathtt{c_{txt}}=\mathtt{E_{txt}}(\mathtt{tp})\).  
Starting from a fully masked sequence \( x_1 \), the generator produces logits
\begin{equation}
l_t=\mathtt{G}(x_t,\alpha_t,\mathtt{c_{txt}}), \qquad x_{t-\frac{1}{T}}= \mathtt{S}(l_t, x_t, t),
\end{equation}
for $k=1, \frac{T-1}{T}, \dots \frac{1}{T}$. After \( T \) steps we obtain visual tokens \( x_0 \), which the image decoder
\( \mathtt{D_{img}} \) converts to a pixel-space image \( I=\mathtt{D_{img}}(x_0) \).

\noindent
\textbf{(ii) Image\,$\boldsymbol{\to}$\,text.}
For captioning, an input image \( I\!\in\!\mathcal{I} \) is tokenized by the image encoder \(\mathtt{E_{img}}\): \(\mathtt{c_{img}} = \mathtt{E_{img}}(I)\). The generator now conditions on the \emph{visual} tokens while progressively decoding text:
\begin{equation}
l_t=\mathtt{G}(x_t, \alpha_t, \mathtt{c_{img}}), \qquad x_{t-\frac{1}{T}}= \mathtt{S}(l_t, x_t, t),
\end{equation}
yielding a text token sequence \( x_0 \), which \( \mathtt{D_{txt}} \) maps to a \( \mathtt{caption}=\mathtt{Detokenize}(\mathtt{D_{txt}}(x_0)) \).

\medskip
\noindent
\textbf{(iii) Image\,+\,question\,$\boldsymbol{\to}$\,answer (VQA).}  
For visual–question answering we supply \emph{both} an image and a question: \( \mathtt{c_{img}}=\mathtt{E_{img}}(I) \) and \( \mathtt{c_{txt}}=\mathtt{E_{txt}}(q) \). They are concatenated and fed to the generator, which outputs logits over answer tokens \( x_k \):
\begin{equation}
l_t=\mathtt{G}(x_t, \alpha_t, [\mathtt{c_{img}}, \mathtt{c_{txt}}]), \qquad x_{t-\frac{1}{T}}= \mathtt{S}(l_t, x_t, t),
\end{equation}
until the full answer \( a \) is produced and decoded by \( a=\mathtt{Detokenize}(\mathtt{D_{txt}}(x_0)) \).

\medskip
\noindent\textbf{Classifier-free guidance.}  
At each decoding step, we apply the same guidance rule, independent of modality:
\begin{equation}
l_k \;\leftarrow\;
\mathtt{G}(z_k, \alpha_k, \mathbf{c})\; + \;\lambda\!\bigl[\mathtt{G}(z_k, \alpha_k, \mathbf{c})-\mathtt{G}(z_k,\alpha_k, \mathbf{c}_{\text{neg}})\bigr],
\end{equation}
where \( z_k \) (image or text tokens) is the partial target sequence, \( \mathbf{c} \) is the \emph{positive} condition (prompt, image, or image\,+question), \( \mathbf{c}_{\text{neg}} \) is the corresponding negative condition, and \( \lambda \) is the guidance scale. Because the loss, decoding schedule, and guidance operator are \emph{identical} in all three scenarios—only the conditioning signal changes—our framework realises a genuinely unified multimodal generator.
\section{Experiment}
\label{sec:exp}

\subsection{Experimental Setup}

\noindent
\textbf{Implementation details.}
We build \name{} on top of the open-source Meissonic models~\citep{bai2024meissonic}.
The MM-DiT backbone is initialized with pretrained weights, and a lightweight linear head is added as a text decoder.
Following Meissonic, we adopt the CLIP~\citep{clip} as text encoder and VQ-VAE as image encoder and decoder, keeping them entirely frozen throughout all experiments.
To support discrete denoising, we append a special \texttt{<mask>} token to CLIP’s vocabulary for text masking, while the image mask token is inherited directly from Meissonic’s initialization.
We observe that, even without training, the \texttt{<mask>} embedding can already be predicted into a coherent sentence during training. Therefore, for simplicity, we freeze the \texttt{<mask>} embedding.
During training, we use a constant learning rate of $1 \times 10^{-4}$ and a weight decay of $1 \times 10^{-2}$. Gradient accumulation is applied in both pretraining and supervised fine-tuning, resulting in an effective batch size of 1024.
We train on 16 H100 GPUs for 5 days.
During inference,  we adopt the default Meissonic configuration, using cosine masking scheduling, 64 sampling steps, and a classifier-free guidance (CFG) scale of 9.0 and 1.5 for text-to-image and image-to-text generation, respectively.

\noindent
\textbf{Training data.}
We train \name{} in two stages using a combination of publicly available and internal datasets, including JourneyDB~\citep{journeydb}, LAION-Art~\citep{laionart}, CC12M~\citep{cc12m}, and others. The final dataset is filtered based on aesthetic score, resolution, and aspect ratio, resulting in approximately 10 million image–text pairs. Both stages are optimized with the unified training objective defined in Eq.~\ref{eq:unified_nelbo}. Below, we describe the datasets and settings for each stage in detail.

\begin{enumerate}[leftmargin=1.5em,label=\textbf{\arabic*.}]
    \item \textbf{Pretraining.} We pretrain \name{} for 100K steps with a batch size of 1024, using the unified objective across both modalities. Text inputs are truncated to a maximum of 77 tokens, and images are resized to $512 \times 512$. The pretraining corpus consists of 8 million image–text pairs, re-captioned using Qwen2.5-VL-3B for improved consistency. Each batch is evenly split between text-to-image and image-to-text samples to enable joint training in both directions.
    \item \textbf{Instruction tuning.} After pretraining, we fine-tune the model on a combination of 1 million instruction-following samples, including LLaVA-Instruct-150K, ALLaVA, SA-1B, and the VQAv2 training set. During this stage, only the answer portion of each prompt is masked. Additionally, we construct a curated dataset of 1 million high-quality image–text pairs to support multi-task training on VQA and image generation. Following the task instructions embedded in each sample, \name{} learns to produce long-form answers, concise replies, and image captions via task-specific prompting.
\end{enumerate}

We present both quantitative and qualitative results for the T2I and I2T tasks in the following sections. Additional experiments and ablation studies are provided in the Appendix.

\subsection{Text-to-Image Generation}

\textbf{Quantitative results.}
Following prior work, we evaluate our $512 \times 512$ model on GenEval~\citep{ghosh2024geneval} after supervised fine-tuning in Tab.~\ref{table:geneval}. \name{} attains an overall accuracy of 0.61, surpassing prior discrete diffusion models such as Monetico (0.44) and Meissonic (0.54), and closely matching Stable Diffusion 3 (0.62) with only 1B parameters. It further shows strong compositional reasoning (0.72 on “Two Objects”, 0.54 on “Counting”), and benefits from joint multimodal training, which enhances T2I performance. These results demonstrate the effectiveness of \name{} as the first unified discrete diffusion model for both text and image modalities.

\begin{figure}[t]
    \vspace{-10pt}
    \centering
    \includegraphics[width=.95\linewidth]{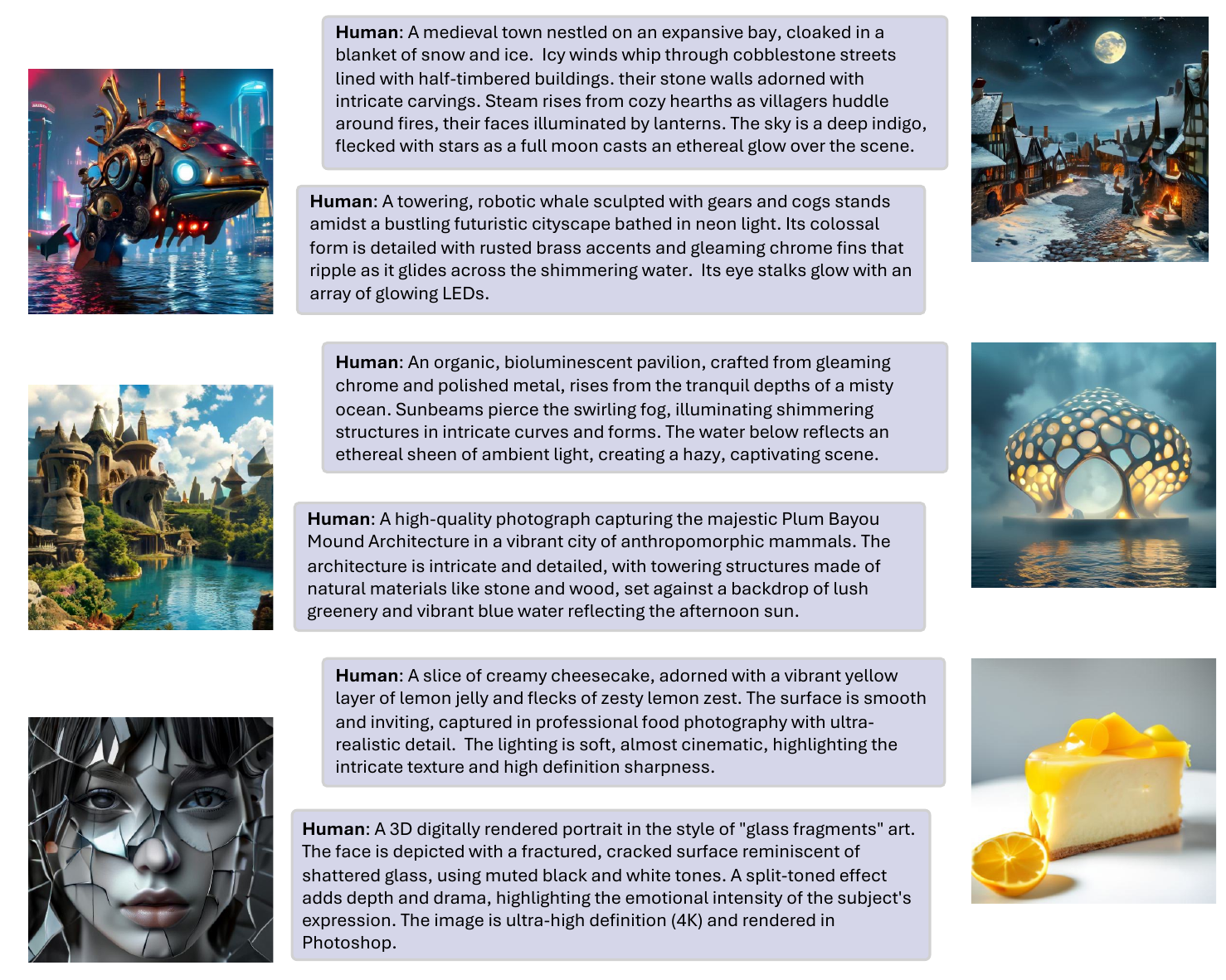}
    \vspace{-5pt}
    \caption{Samples of Text-to-Image Generation by \name{}.}
    \label{fig:t2i_1}
\end{figure}

\begin{figure}[t]
    \vspace{-10pt}
    \centering
    \includegraphics[width=.95\linewidth]{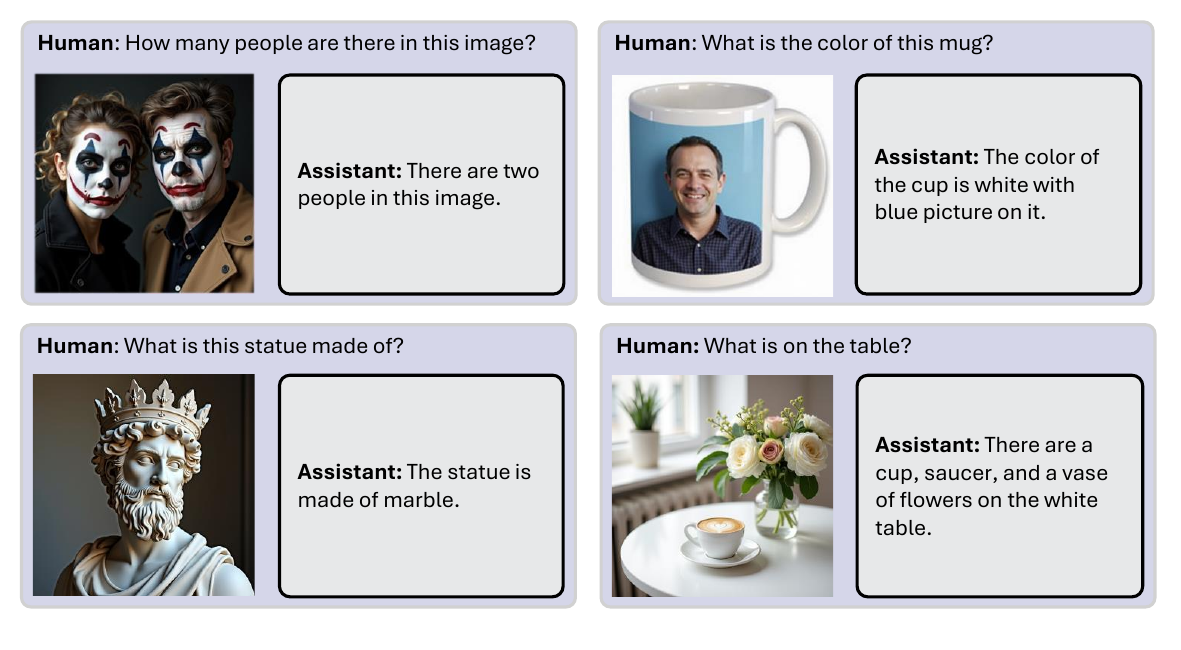}
    \vspace{-5pt}
    \caption{Samples of Visual Question Answering by \name{}.}
    \label{fig:vqa_v2}
    \vspace{-10pt}
\end{figure}

\begin{table}[t]
\caption{Evaluation of text-to-image generation performance on the GenEval~\citep{ghosh2024geneval}.}\label{table:geneval}
\centering
\vspace{-5pt}
\resizebox{\linewidth}{!}{
\setlength{\tabcolsep}{0.75pt}
\begin{tabular}{lcccccccccc} 
\cmidrule[\heavyrulewidth]{1-4}\cmidrule[\heavyrulewidth]{5-11}
\multirow{2}{*}{\textbf{Model}} & \multirow{2}{*}{\begin{tabular}[c]{@{}c@{}}\textbf{Text Gen} \\Arch\end{tabular}} & \multirow{2}{*}{\begin{tabular}[c]{@{}c@{}}\textbf{Image Gen} \\Arch\end{tabular}} & \multirow{2}{*}{\begin{tabular}[c]{@{}c@{}}\textbf{Params} \\(B)\end{tabular}} & \multirow{2}{*}{\textbf{Overall~$\uparrow$}} & \multicolumn{2}{c}{\textbf{Objects}~$\uparrow$} & \multirow{2}{*}{\textbf{Counting}~$\uparrow$} & \multirow{2}{*}{\textbf{Colors}~$\uparrow$} & \multirow{2}{*}{\textbf{Position}~$\uparrow$} & \multirow{2}{*}{\begin{tabular}[c]{@{}c@{}}\textbf{Color}~$\uparrow$ \\\textbf{Attribution}\end{tabular}} \\ 
\cmidrule{6-7}
 &  &  &  &  & \small \textbf{Single} & \small \textbf{Two} &  &  &  &  \\ 
\midrule
PixArt-$\alpha$~\citep{pixart} & - & Diffusion & 0.6 & 0.48 & 0.98 & 0.50 & 0.44 & 0.80 & 0.08 & 0.07 \\
SD 2.1~\citep{stable_diffusion} & - & Diffusion & 0.9 & 0.50 & 0.98 & 0.51 & 0.44 & 0.85 & 0.07 & 0.17 \\
DALL-E 2~\citep{ramesh2022hierarchical} & - & Diffusion & 6.5 & 0.52 & 0.94 & 0.66 & 0.49 & 0.77 & 0.10 & 0.19 \\
SDXL~\citep{podell2023sdxl} & - & Diffusion & 2.6 & 0.55 & 0.98 & 0.74 & 0.39 & 0.85 & 0.15 & 0.23 \\
DALL-E 3~\citep{dalle3} & - & Diffusion & - & 0.67 & 0.96 & 0.87 & 0.47 & 0.83 & 0.43 & 0.45 \\ 
SD 3 \citep{esser2024scaling} & - & Diffusion & 2 & 0.62 & 0.98 & 0.74 & 0.63 & 0.67 & 0.34 & 0.36 \\ 
\midrule
LWM~\citep{lwm} & AR & AR & 7 & 0.47 & 0.93 & 0.41 & 0.46 & 0.79 & 0.09 & 0.15 \\
SEED-X~\citep{seedx} & AR & AR & 17 & 0.49 & 0.97 & 0.58 & 0.26 & 0.80 & 0.19 & 0.14 \\
Chameleon~\citep{team2024chameleon} & AR & AR & 7 & 0.39 & - & - & - & - & - & - \\
Show-O~\citep{showo} & AR & Discrete Diff. & 1.3 & 0.68 & 0.98 & 0.80 & 0.66 & 0.84 & 0.31 & 0.50 \\ 
Transfusion~\citep{zhou2024transfusion} & AR & Diffusion & 8 & 0.67 & - & - & - & - & - & - \\
D-DiT~\citep{ddit} & Discrete Diff. & Diffusion & 2 & 0.65 & 0.97 & 0.80 & 0.54 & 0.76 & 0.32 & 0.50 \\
\midrule
Monetico ($512\times512$)~\citep{bai2024meissonic} & - & Discrete Diff. & 1 & 0.44 & 0.92 & 0.48 & 0.26 & 0.78 & 0.06 & 0.13 \\
Meissonic ($1024\times1024$)~\citep{bai2024meissonic} & - & Discrete Diff. & 1 & 0.54 & 0.99 & 0.66 & 0.42 & 0.86 & 0.10 & 0.22 \\
UniDisc ($512\times 512$)~\citep{unidisc} & Discrete Diff. & Discrete Diff. & 1.4 & 0.42 & 0.92 & 0.47 & 0.15 & 0.67 & 0.13 & 0.19 \\
\midrule
\name{} ($512\times 512$) & Discrete Diff. & Discrete Diff. & 1 & 0.61 & 0.98 & 0.72 & 0.54 & 0.82 & 0.19 & 0.41 \\
\bottomrule
\end{tabular}}
\end{table}

\textbf{Qualitative results.}
We present diverse generations from our model conditioned on rich textual prompts in Fig.~\ref{fig:t2i_1}. The outputs exhibit strong text-image alignment, capturing fine details in both realistic and imaginative scenes. Our model effectively renders complex structures, lighting, and textures across various domains.

\subsection{Image-to-Text Generation}

We present a comprehensive comparison of our model \name{} against other multimodal models across four benchmarks: MS-COCO (image captioning)~\citep{coco}, VQAv2~\citep{vqa}, MME~\citep{mme}, GQA~\citep{gqa}, and MMMU~\citep{mmmu} in Tab.~\ref{table:lm_benchmarks}. Notably, \name{} is the first unified model to employ discrete diffusion for both text-to-image and image-to-text generation, demonstrating that this approach is highly competitive.

\textbf{Quantitative comparison.}
\name{} demonstrates strong performance across benchmarks despite its compact size (1B parameters). It achieves a CIDEr score of 59.9 on MS-COCO, surpassing diffusion-based baselines such as D-DiT (56.2). On VQAv2, it attains 68.2\% accuracy, outperforming Show-O and D-DiT. Moreover, it reaches 1107.4 on MME and 57.5 on GQA, underscoring its competitiveness across multimodal reasoning tasks. These results highlight the effectiveness of \name{} as a unified diffusion-based model that balances efficiency with high-quality task performance.

\begin{figure}[t]
    \centering
    \includegraphics[width=.95\linewidth]{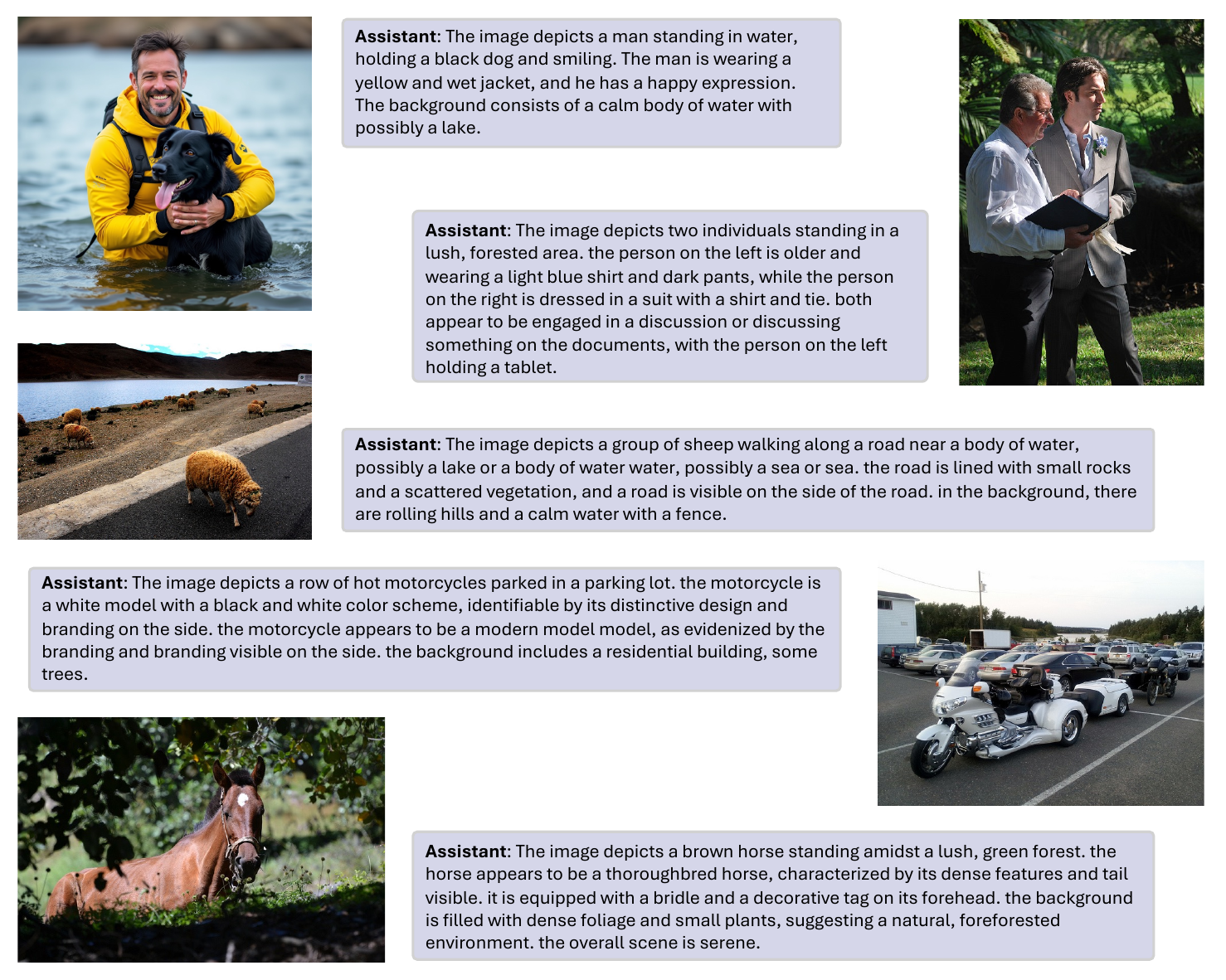}
    \vspace{-5pt}
    \caption{Samples of Image-to-Text Generation by \name{}.}
    \label{fig:i2t}
    \vspace{-10pt}
\end{figure}

\begin{table*}[ht]
    \caption{Evaluation of image captioning, visual question answering on multimodal benchmarks.}
    \label{table:lm_benchmarks}
    \vspace{-5pt}
    \centering
    \resizebox{\linewidth}{!}{
        \begin{tabular}{lcccccccc}
            \toprule
            \textbf{Model} 
            & \shortstack[c]{\strut \textbf{Params} \\ \strut \textbf{(B)}} 
            & \shortstack[c]{\strut \textbf{Text Gen} \\ \strut \textbf{Arch}} 
            & \shortstack[c]{\strut \textbf{Image Gen} \\ \strut \textbf{Arch}} 
            & \shortstack[c]{\strut \textbf{MS-COCO} \\ \strut \textbf{CIDEr~$\uparrow$}} 
            & \shortstack[c]{\strut \textbf{VQAv2} \\ \strut \textbf{Acc.~$\uparrow$}} 
            & \shortstack[c]{\strut \textbf{MME} \\ \strut \textbf{Acc.~$\uparrow$}} 
            & \shortstack[c]{\strut \textbf{GQA} \\ \strut \textbf{Acc.~$\uparrow$}} 
            & \shortstack[c]{\strut \textbf{MMMU} \\ \strut \textbf{Acc.~$\uparrow$}} \\
            \midrule
            InternVL-2.0~\citep{chen2025internvl2} & 8 & AR & - & - & - & 1648.1 & 61.0 & 49.3 \\
            LLaVA-Next~\citep{liu2024llavanext} & 13 & AR & - & - & 82.8 & 1575.0 & 65.4 & 36.2 \\
            BLIP-2~\citep{blip2} & 13 & AR & - & - & 65.0 & 1293.8 & 41.0 & 34.4 \\
            QWEN-VL~\citep{bai2023qwenvl} & 7 & AR & - & - & 78.2 & 1487.5 & 57.5 & 35.9 \\
            OpenFlamingo~\citep{awadalla2023openflamingo} & 9 & AR & - & 65.5 & 43.5 & - & - & 28.7 \\
            Flamingo~\citep{flamingo} & 9 & AR & - & 79.4 & 51.8 & - & - & - \\
            \midrule
            Chameleon~\citep{team2024chameleon} & 7 & AR & AR & 18.0 & - & - & - & - \\
            LWM~\citep{lwm} & 7 & AR & AR & - & 55.8 & - & - & - \\
            Show-O (256$\times$256)~\citep{showo} & 1.3 & AR & Discrete Diff. & - & 64.7 & 1014.9 & 54.2 & - \\
            Show-O (512$\times$512)~\citep{showo} & 1.3 & AR & Discrete Diff. & - & 69.4 & 1097.2 & 58.0 & 27.4 \\
            Transfusion~\citep{zhou2024transfusion} & 7 & AR & Diffusion & 29.0 & - & - & - & - \\
            D-DiT (256$\times$256)~\citep{ddit} & 2 & Discrete Diff. & Diffusion & - & 59.5 & 897.5 & 55.1 & - \\
            D-DiT (512$\times$512)~\citep{ddit} & 2 & Discrete Diff. & Diffusion & 56.2 & 60.1 & 1124.7 & 59.2 & - \\
            UniDisc~\citep{unidisc} & 1.4 & Discrete Diff. & Discrete Diff. & 46.8 & - & - & - & - \\
            \midrule
            \name{} (512$\times$512) & 1 & Discrete Diff. & Discrete Diff. & 59.9 & 68.2 & 1107.4 & 57.5 & 27.6 \\
            \name{} (1024$\times$1024) & 1 & Discrete Diff. & Discrete Diff. & 60.1 & 70.2 & 1139.2 & 57.8 & 28.7 \\
            \bottomrule
        \end{tabular}
    }
\end{table*}

\textbf{Qualitative results.} We present example captions generated by our model across diverse scenarios in Fig.~\ref{fig:i2t}, including humans, animals, vehicles, and natural landscapes. The model demonstrates strong visual grounding and fine-grained descriptive ability, accurately capturing attributes such as clothing, expressions, background context, and object relationships.
Fig.~\ref{fig:vqa_v2} illustrates our model’s ability to accurately answer visual questions across various domains, including object counting, color recognition, material identification, and compositional reasoning.


\subsection{Ablation Study and Analysis}

\noindent
\textbf{Analysis of the inference timesteps.} 
As shown in Tab.~\ref{tab:diffusion_timesteps}, performance generally improves with more diffusion steps, plateauing around $T = 32$. GenEval and CIDEr see large gains from $T = 8$ to $T = 32$, with diminishing returns afterward. VQAv2 remains stable across timesteps, indicating that fewer steps suffice for discriminative tasks. Overall, a moderate number of steps provides a good balance between accuracy and efficiency.

\begin{table}[t]
  \centering
  \begin{minipage}[t]{0.48\textwidth}
      \caption{Impact of text loss weight. We apply the same text loss weight during both pretraining and instruction tuning.}
      \label{tab:text_loss_weight}
      \vspace{-5pt}
      \centering
      \scriptsize          
      \begin{tabular}{lccccc}
          \toprule
          \textbf{Benchmark} & \textbf{0.2} & \textbf{0.4} & \textbf{0.6} & \textbf{0.8} & \textbf{1.0}\\
          \midrule
          GenEval        & 60.1 & 60.5 & \textbf{61.6} & 60.8 & 58.3 \\
          MS-COCO        & 51.4 & 52.1 & \textbf{59.9} & 58.8 & 59.4 \\
          VQAv2          & 62.7 & 66.2 & 68.2 & 68.4 & \textbf{69.2} \\
          \bottomrule
       \end{tabular}
  \end{minipage}
  \hfill
  \begin{minipage}[t]{0.48\textwidth}
    \caption{Effect of joint training. We denote text-to-image as T2I and image-to-text as I2T, respectively.}
    \label{tab:joint_training}
    \vspace{-5pt}
    \centering
    \scriptsize
    \begin{tabular}{lccc}
      \toprule
      \textbf{Benchmark} & \textbf{T2I only} & \textbf{I2T only} & \textbf{Joint training} \\
      \midrule
      GenEval       & 59.3 & 28.3 & \textbf{61.6} \\
      MS-COCO       & -    & \textbf{60.1} & 59.9 \\
      VQAv2         & -    & \textbf{69.1} & 68.2 \\
      \bottomrule
    \end{tabular}
  \end{minipage}
  \vspace{-10pt}
\end{table}

\begin{wraptable}{r}{0.48\linewidth}
  \centering
  \caption{Performance across different diffusion timesteps.}
  \label{tab:diffusion_timesteps}
  \resizebox{\linewidth}{!}{
    \begin{tabular}{lccc}
        \toprule
        \textbf{Sample steps} & \textbf{GenEval} & \textbf{CIDEr} & \textbf{VQAv2} \\
        \midrule
        T=8   & 51.6 & 43.6 & 53.9 \\
        T=16  & 58.5 & 59.3 & 57.4 \\
        T=24  & 59.3 & 59.4 & 62.3 \\
        T=32  & \textbf{61.9} & 59.7 & 65.4 \\
        T=40  & 61.7 & 60.1 & 66.8 \\
        T=64  & 61.1 & 59.9 & \textbf{68.2} \\
        \bottomrule
    \end{tabular}
  }
\end{wraptable}

\noindent
\textbf{Analysis of the text loss weight.} 
As shown in Tab.~\ref{tab:text_loss_weight}, moderate text loss weights (around 0.6) yield the best overall performance. CIDEr and GenEval peak near this value, suggesting that both insufficient and excessive text weighting can harm generation quality. VQAv2 continues to improve with stronger text supervision but begins to plateau beyond 0.6. Overall, while discriminative tasks benefit from heavier textual guidance, generative tasks require a balanced mix of visual and textual signals—highlighting the importance of grounding language in multimodal learning.

\noindent
\textbf{Analysis of joint training.} 
Joint optimization over both text-to-image (T2I) and image-to-text (I2T) objectives is essential. As shown in Tab.~\ref{tab:joint_training}, joint training yields the highest GenEval score, outperforming both T2I-only and I2T-only variants. Notably, I2T-only causes GenEval to drop sharply from 61.6 to 28.3—more than a twofold decrease—while MS-COCO CIDEr remains nearly unchanged and VQAv2 declines only slightly. These results show that separating the objectives severely weakens cross-modal integration, underscoring the need for unified optimization to maintain strong multimodal coherence.

\subsection{The scalability of Muddit}

To demonstrate the scalability of our approach, we curate roughly 10 million image–text pairs from LAION-ART~\citep{laionart}, JourneyDB~\citep{journeydb}, CC12M~\citep{cc12m} and internal datasets. We filter out samples with an aesthetic score below 7, a height or width under 512 pixels, or an aspect ratio above 2. All images are re-captioned using Qwen2.5-VL 7B~\citep{bai2023qwenvl}.
We pretrain Muddit on this dataset with a batch size of 512 and a resolution of 1024, applying random masking to both image and text modalities. The image and text loss weights are set to 1.0 and 0.3, respectively. Training runs for 100K steps. 

For instruction tuning, we collect about 6M samples from LLAVA-Instruct-150K~\citep{llava}, ALLaVA LAION~\citep{allava}, SA-1B~\citep{kirillov2023sam}, ART500K~\citep{art500k}, ScienceQA~\citep{scienceqa}, Chart2Text~\citep{chart2text}, and VQAv2~\citep{vqa}. Muddit is then trained with a batch size of 512 at a resolution of 1024, with masking applied only to the answer text. We also add a 2M high-quality image dataset for high-quality fine-tuning.
Further training configurations are provided in Tab.~\ref{tab:scaling_training_config}. All experiments are conducted on 16 H100 GPUs.

We evaluate the scaled Muddit model against other comparably sized unified models and state-of-the-art unified discrete diffusion models~\citep{luminadimoo, mmada}, as shown in Tab.~\ref{tab:scale_quan_comp}. Across established benchmarks, Muddit exhibits consistent improvements in both image generation and image understanding, empirically validating the scalability of our model. Furthermore, we compare Muddit with unified models of similar parameter sizes, all of which rely on hybrid architectures. Despite being trained on substantially less data, Muddit achieves superior performance.

We attribute this data efficiency to two key factors. First, our visual prior naturally maintains strong text-following capability for text-to-image generation, enabling robust alignment between image and text modalities. From the perspective of unified modeling, we prioritize cross-modal alignment over isolated single-modality ability, which allows Muddit to reach higher performance with less training data. Second, Muddit adopts a fully unified modeling paradigm: the model learns by predicting mask tokens based on context across all tasks (text-to-image and image-to-text). In contrast, hybrid architectures must simultaneously handle next-token prediction alongside velocity or mask prediction, and often introduce additional special tokens (\textit{e.g.}, $<\text{soi}>$, $<\text{eoi}>$), which increases architectural complexity and hinders optimization.

\begin{table*}[t]
\centering
\caption{Training hyperparameters across different training stages.}
\label{tab:scaling_training_config}
\vspace{-5pt}
\resizebox{0.6\linewidth}{!}{
\begin{tabular}{lcc}
\toprule
\textbf{Hyperparameters} & \textbf{Stage-I} (Pre-training) & \textbf{Stage-II} (Instruction-tuning) \\
\midrule
Learning Rate & $1.0\times10^{-4}$ & $1.0\times10^{-4}$ \\
LR Scheduler & Constant & Constant \\
Weight Decay & 0.01 & 0.01 \\
Max Gradient Norm & 10.0 & 10.0 \\
Optimizer & \multicolumn{2}{c}{AdamW ($\beta_1=0.9,\ \beta_2=0.999$)} \\
Batch Size & 512 & 512 \\
Training Steps & 100K & 15K \\
Training GPUs & 16$\times$H100 & 16$\times$H100 \\
\midrule

Gen. Resolution & 1024 & 1024 \\

Under. Resolution & 1024 & 1024 \\
\bottomrule
\end{tabular}
}
\end{table*}

\begin{table}[t]
\centering
\caption{Quantitative comparison with other unified models.}
\label{tab:scale_quan_comp}
\resizebox{\linewidth}{!}{
\begin{tabular}{lcccccccc}
\toprule
\textbf{Model} & \textbf{Params} & \textbf{Base model} & \textbf{Architecture} & \textbf{Data scale} & \textbf{Geneval w/ TTS} & \textbf{VQAv2} & \textbf{MME} & \textbf{MMMU} \\
\midrule
Lumina-DiMOO~\citep{luminadimoo}                & 8B   & LLaDA & Discrete Diff. & 80M     & 0.92 & --   & 1534.2 & 58.6 \\ 
MMaDA ($512 \times 512$)~\citep{mmada}    & 8B   & LLaDA  & Discrete Diff. & Unknown & 0.66 & 76.7 & 1410.7 & 30.2 \\ 
\midrule
Show-O (512$\times$512)~\citep{showo} & 1.3B & Phi-1.5 & AR + Discrete Diff & 35M & -- & 69.4 & 1097.2 & 27.4 \\ 
D-DiT (512$\times$512)~\citep{ddit} & 2B & SD3-medium & Discrete Diff. + Diff. & 40M & -- & 60.1 & 1124.7 & -- \\
Muddit (512$\times$512) & 1B & Meissonic & Discrete Diff. & 10M & 0.64 & 68.2 & 1107.4 & 27.6 \\
Muddit (1024$\times$1024) & 1B & Meissonic & Discrete Diff & 16M & 0.67 & 70.2 & 1139.2 & 28.7 \\ 
\bottomrule
\end{tabular}
}
\end{table}

\section{Conclusion}
\label{sec:conclusion}
In this work, we present the second-generation Meissonic: \name{}, a unified generative framework that employs discrete diffusion to bridge text and image modalities. By unifying image and text generation within a single model, \name{} demonstrates strong performance across text-to-image, image-to-text, and VQA tasks. Notably, it matches or outperforms the capabilities of significantly larger autoregressive models, while enabling fast, parallel inference. Our results validate the effectiveness of discrete diffusion as a general-purpose modeling strategy and highlight its potential to serve as a scalable backbone for future multimodal systems that are equipped with strong visual priors.

\noindent
\textbf{Acknowledgement.} This work is supported by the National Key Research and Development Program of China (No. 2023YFC3807600) and the National Natural Science Foundation of China under Grant No. 62320106007. In addition, this work is supported in part by NUS Start-up Grant A-0010106-00-00.

\clearpage

\bibliography{main}
\bibliographystyle{main}

\clearpage

\appendix

\section*{\LARGE Appendix}
\addcontentsline{toc}{section}{Appendix}

\section*{Appendix Overview}

This appendix provides additional discussions, results, and analyses to complement the main paper. 
It is organized as follows:

\begin{itemize}
    \item \textbf{Related Work} (Sec.~\ref{sec:related_work}): 
    We review unified multimodal models for understanding and generation, 
    with a focus on autoregressive and diffusion-based paradigms, as well as recent advances in masked image modeling.

    \item \textbf{Additional Qualitative Results} (Sec.~\ref{sec:qualitative}): 
    We present extended visualizations for several tasks, including image captioning, text-to-image generation, 
    visual question answering, and image-guided text editing.

    \item \textbf{Additional Experimental Results} (Sec.~\ref{sec:additional_exp_res}):
    We present more experimental results.

    \item \textbf{Additional Ablation Studies} (Sec.~\ref{sec:additional_ablation_study}): 
    We present extended ablation studies.

    \item \textbf{Inference Time Analysis} (Sec.~\ref{sec:inference}): 
    We analyze inference efficiency by comparing autoregressive decoding with discrete diffusion, 
    providing FLOPs complexity and speed benchmarks.

    \item \textbf{Generated Results Step by Step} (Sec.~\ref{sec:step}): 
    We illustrate the reverse discrete diffusion process in detail, 
    showing intermediate decoding steps and examples of progressive generation.

    \item \textbf{Discussion} (Sec.~\ref{sec:discussion}): 
    We reflect on the limitations of our approach and its broader impacts, 
    including potential applications and risks of misuse.

    \item \textbf{Use of Large Language Models} (Sec.~\ref{sec:usage_llm}): 
    We clarify the role of large language models during paper preparation.
\end{itemize}

\section{Related Work}
\label{sec:related_work}

\subsection{Unified Models For Generation and Understanding}

The success of LLMs in language modeling has inspired efforts to extend unified generation to multimodal domains. However, the divergence between autoregressive and diffusion-based paradigms presents fundamental architectural trade-offs.
Autoregressive models naturally handle language, and several works~\citep{sun2023generative, wang2024illume, tong2024metamorph, seedx, dong2023dreamllm, chen2025multimodal} extend this by connecting vision modules to LLMs via adapters or instruction tuning, with LLMs serving as planning modules that produce intermediate representations for image generation. 
While effective to some extent, these paradigms often exhibit limited interaction between text and image modalities and struggle with content consistency, particularly in image-to-image generation and complex instruction-based synthesis.
To address these limitations, recent research explores unified generation models that integrate understanding and generation within a single architecture. We categorize these into four major paradigms (see Fig.~\ref{fig:unified_cls}):

\textbf{Fully Autoregressive}: Both text and image are tokenized into discrete sequences and modeled with an AR Transformer~\citep{liu2024world, team2024chameleon, wu2024vila, wang2024emu3, janus-pro, lumina-mgpt, vsa, adapt}. These models achieve strong cross-modal generation but suffer from high latency due to sequential decoding.

\textbf{Text AR, Image Diffusion}: LLMs generate text tokens while image synthesis is delegated to pretrained continuous diffusion backbones~\citep{zhou2024transfusion, zhao2024monoformer, ma2024janusflow} or discrete diffusion~\citep{showo}. Though visually strong, these models are not truly unified, as they rely on separate architectures and token spaces.

\textbf{Image Diffusion, Text Discrete Diffusion}: Emerging models experiment with discrete diffusion for text and images~\citep{li2024dual}, though many, like Dual-Diffusion~\citep{li2024dual}, still use continuous diffusion for image synthesis, failing to realize true modality symmetry.

\textbf{Fully Discrete Diffusion}: Recent work like UniDisc~\citep{swerdlow2025unified} pioneers full-token discrete diffusion over shared Transformer backbones. These models support parallel sampling and native integration, but currently lag behind in generation fidelity and scale.

Among these, the GPT-4o~\citep{openai2025gpt4o} model represents a significant advance as a unified multimodal generative system. However, its closed-source nature obscures critical architectural and training details, and its success may be largely attributable to scale rather than architectural novelty~\citep{chen2025empirical}.

\subsection{Masked Image Modeling}

Masked Image Modeling (MIM) has emerged as a powerful self-supervised learning paradigm in computer vision~\citep{bai2025masks}, drawing inspiration from the success of Masked Language Modeling (MLM) in NLP, notably BERT~\citep{devlin2018bert}. The fundamental principle of MIM involves obscuring portions of an image, which could be raw pixels (MAE~\citep{he2022masked}), latent patches of pixels, or even discrete latent tokens (BEiT~\citep{bao2021beit}, MaskGIT~\citep{chang2022maskgit}), and training a model, typically an autoencoder, to predict or reconstruct this missing information by leveraging the context provided by the visible parts.

MaskGIT~\citep{chang2022maskgit} introduced parallel decoding via iterative token refinement, inspiring discrete diffusion models. Recent work such as RandomAR~\citep{fan2024fluid} and MAR~\citep{li2024autoregressive} formalize this as random-order or masked autoregressive generation, blending AR and MIM principles. The major conceptual difference between
RandomAR/MAR and MaskGIT is in the scanning order at inference time.

This class of techniques forms the conceptual foundation of discrete diffusion over tokenized spaces~\citep{shi2025rectok} and plays a critical role in modern unified models. We will introduce discrete diffusion in the next section.

\subsection{Relationship to Concurrent Work}
Our main contribution is to demonstrate that a unified, \emph{visual-prior} fully discrete diffusion model can be both effective and data-efficient for image understanding tasks, rather than just text-to-image generation tasks. Regarding the distinction from concurrent discrete diffusion works~\citep{mmada, bai2026prism, luminadimoo, xin2025dmllm}, we think that unified models should allow for multiple design choices. Our goal is to demonstrate that a visual-first, fully discrete diffusion backbone can be a practical and competitive alternative to the more common “LLM-first” unified paradigm, and we believe this is a fundamental design choice.

Specifically, prior unified discrete diffusion models, such as UniDisc~\citep{unidisc}, are trained from scratch on multimodal data and therefore lack strong visual priors. As a result, they significantly underperform early diffusion baselines such as Stable Diffusion 1.5~\citep{stable_diffusion} and do not support visual question answering tasks~\citep{vqa}. In contrast, Muddit is the first unified discrete diffusion model built on top of a pretrained high-resolution text-to-image backbone~\citep{bai2024meissonic}, with a lightweight text decoder on top. 
This visual prior is not an implementation detail: it improves the scalability and generalization behavior of discrete diffusion through a visual backbone that is well aligned with text.

\section{Additional Qualitative Results}
\label{sec:qualitative}
\begin{figure}[t]
    \centering
    \includegraphics[width=.85\linewidth]{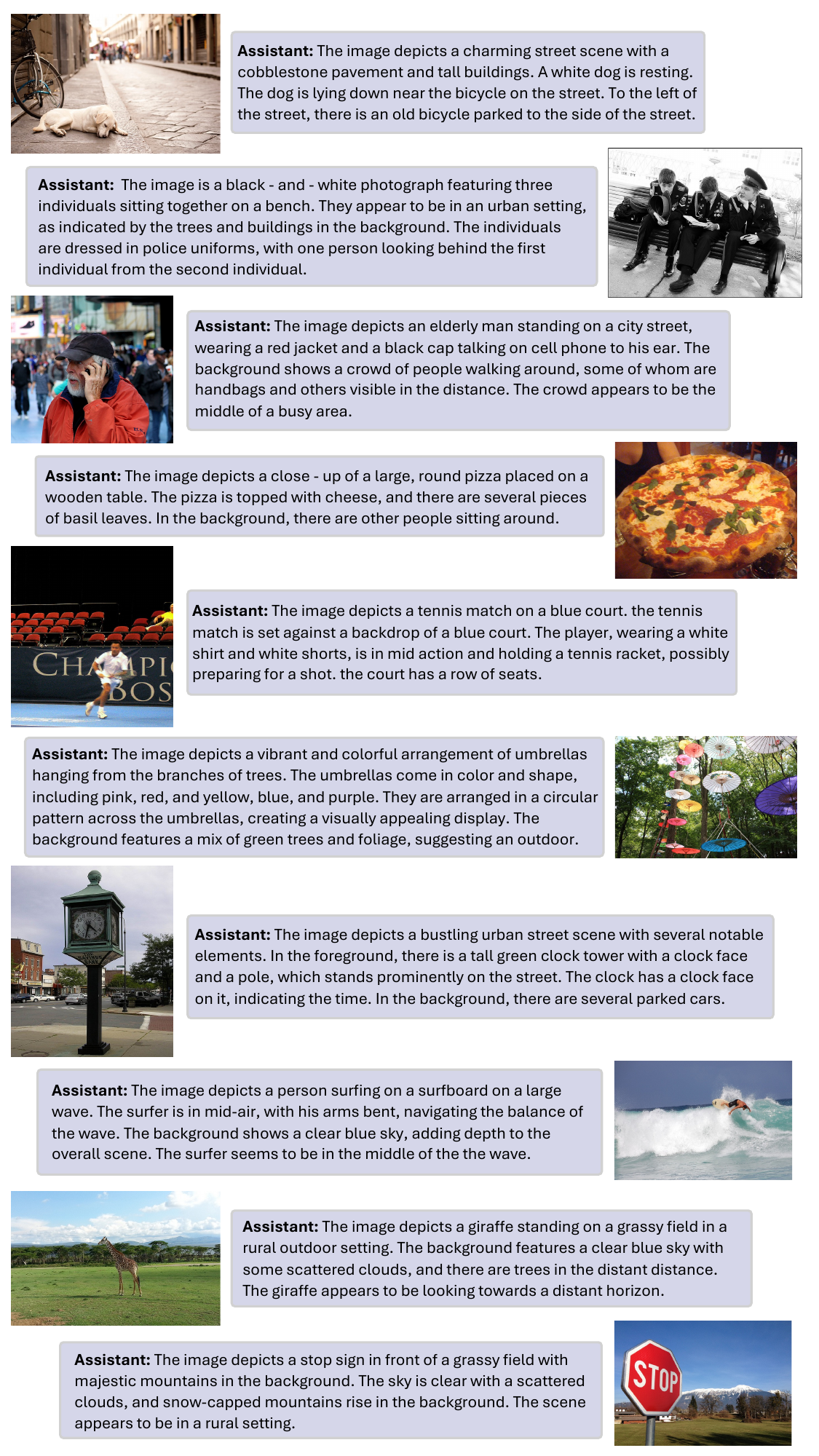}
    \caption{Image-to-text generated results.}
    \label{fig:caption_1}
\end{figure}

\noindent
\textbf{Image-to-text Generation.}
We present more examples for image-to-text generation in Fig.~\ref{fig:caption_1}.

\begin{figure}[t]
    \centering
    \includegraphics[width=1.\linewidth]{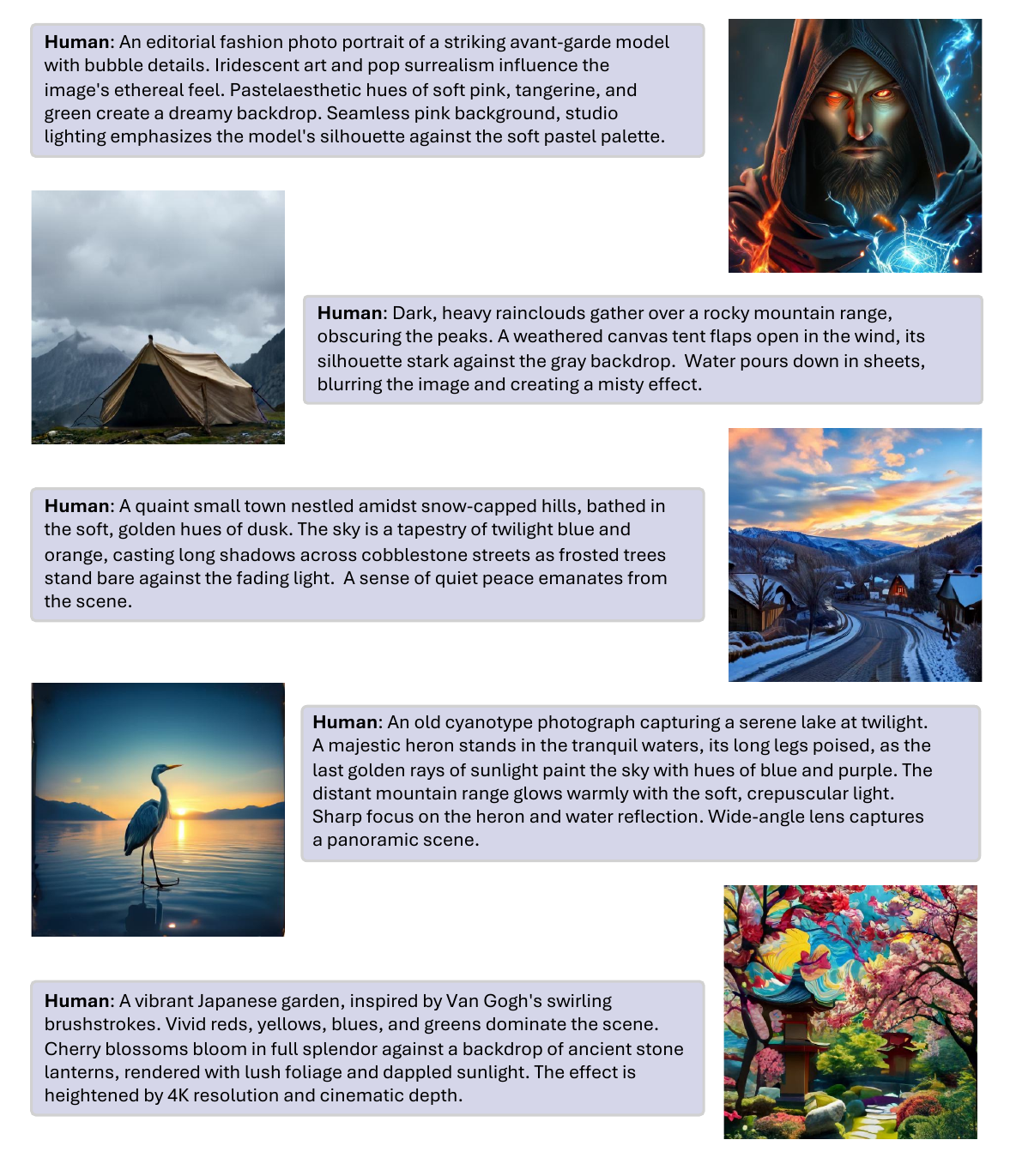}
    \caption{Text-to-image generation results.}
    \label{fig:t2i}
\end{figure}

\noindent
\textbf{Text-to-image Generation.}
We present more examples for text-to-image generation in Fig.~\ref{fig:t2i}.

\begin{figure}[t]
    \centering
    \includegraphics[width=1.\linewidth]{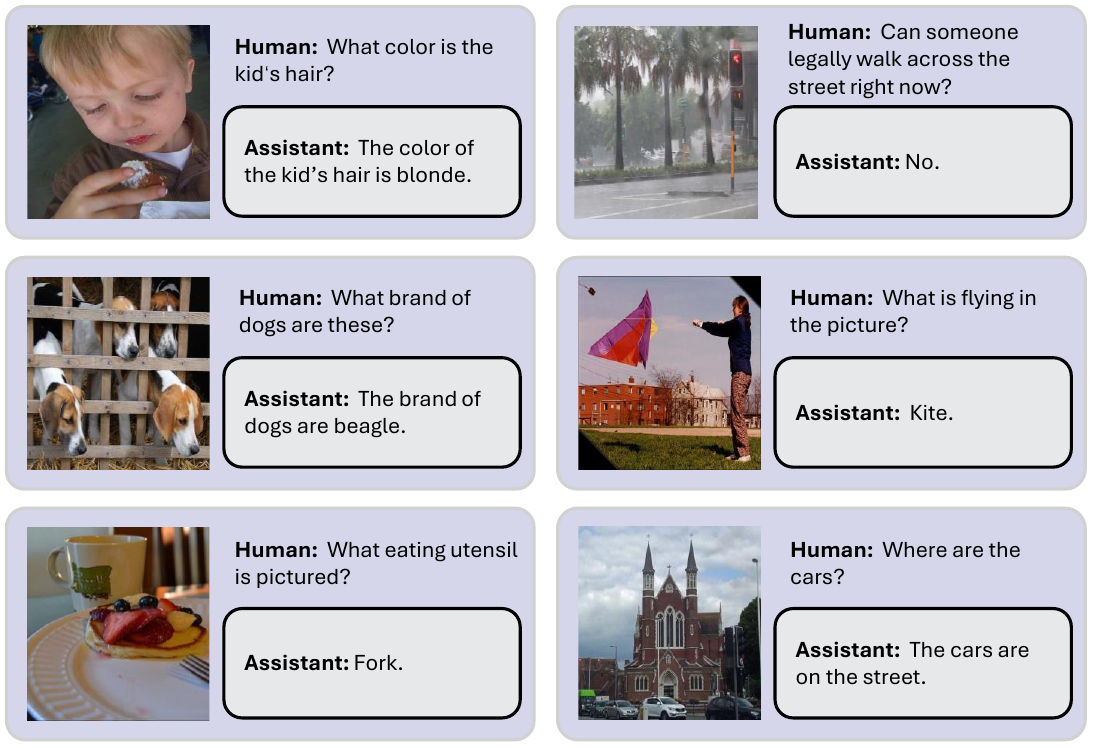}
    \caption{Visual question answering results.}
    \label{fig:vqa}
\end{figure}

\noindent
\textbf{Visual Question Answering.}
We present more examples for visual question answering in Fig.~\ref{fig:vqa}. \name{} reliably identifies fine-grained attributes (\textit{e.g.}, “blonde” hair), object categories (\textit{e.g.}, “beagle”), and physical affordances (\textit{e.g.}, answering “No” to crossing at a red light). Notably, it also handles commonsense reasoning and spatial localization, such as inferring traffic legality or locating vehicles on the street.

\begin{figure}[t]
    \centering
    \includegraphics[width=1.\linewidth]{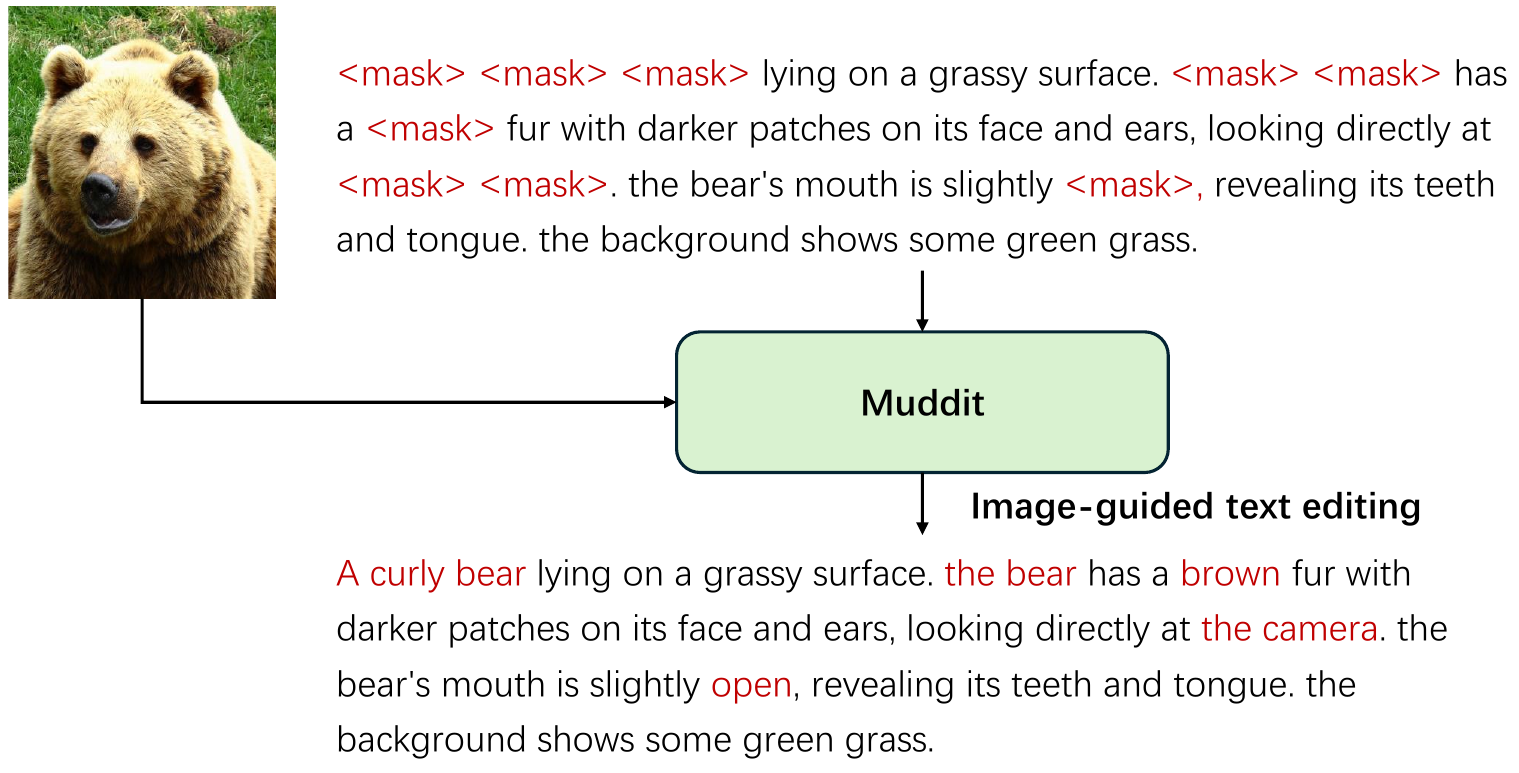}
    \caption{Image-guided text editing results.}
    \label{fig:igte}
\end{figure}

\noindent
\textbf{Image-guided text editing.}
Zero-shot text-guided image editing performance is already verified and presented in Meissonic~\citep{bai2024meissonic}. As the successor to Meissonic, we present \name{}'s performance on the image-guided text editing task, where the model completes a masked sentence based on the input image. As shown in Fig.~\ref{fig:igte}, given a partially masked caption and an image, \name{} fills in the blanks with semantically and visually grounded phrases. 

\begin{figure}[t]
    \centering
    \includegraphics[width=1.\linewidth]{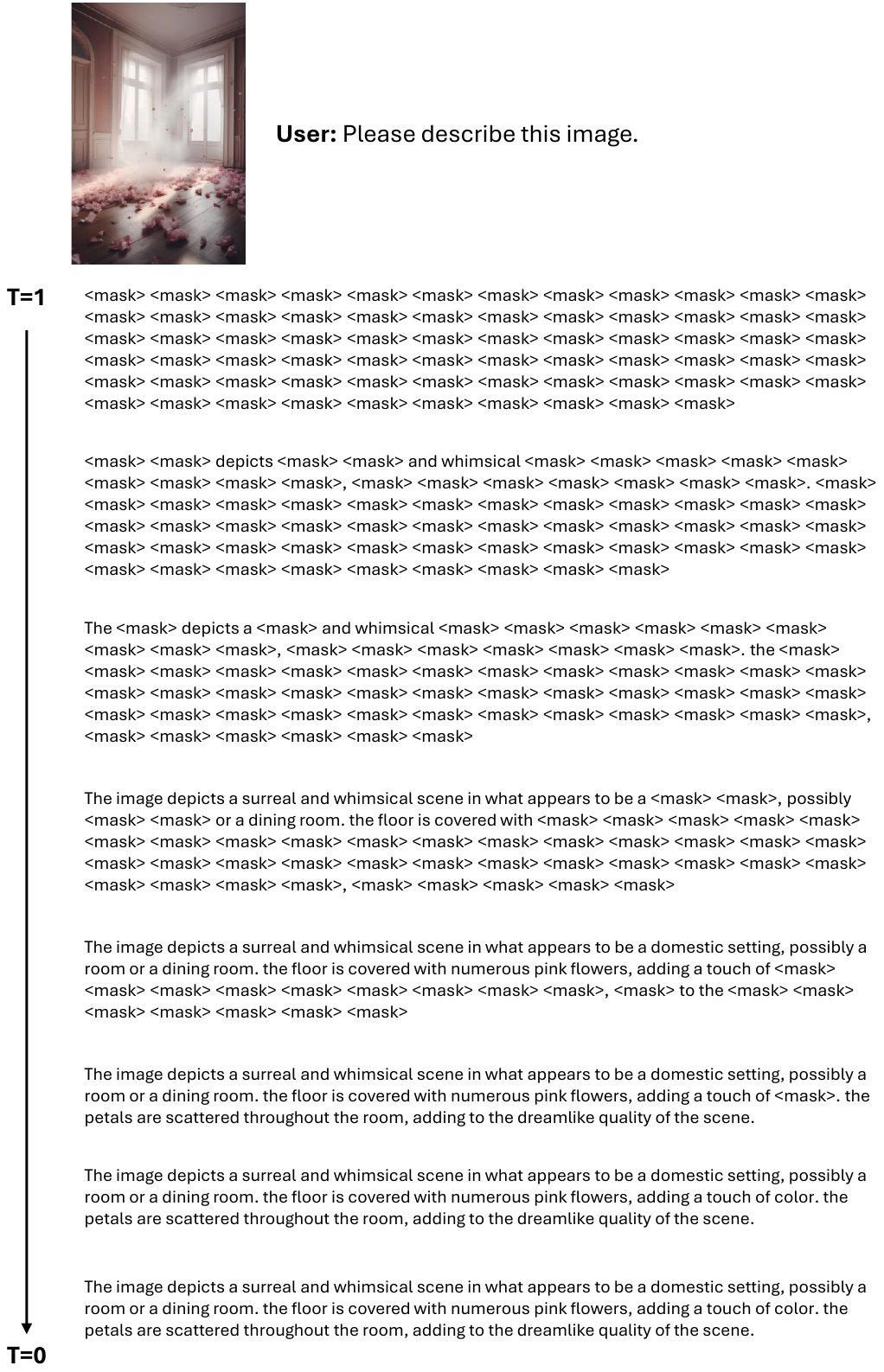}
    \caption{Image-to-text generated results in each step.}
    \label{fig:room_sbs}
\end{figure}

\begin{figure}[t]
    \centering
    \includegraphics[width=1.\linewidth]{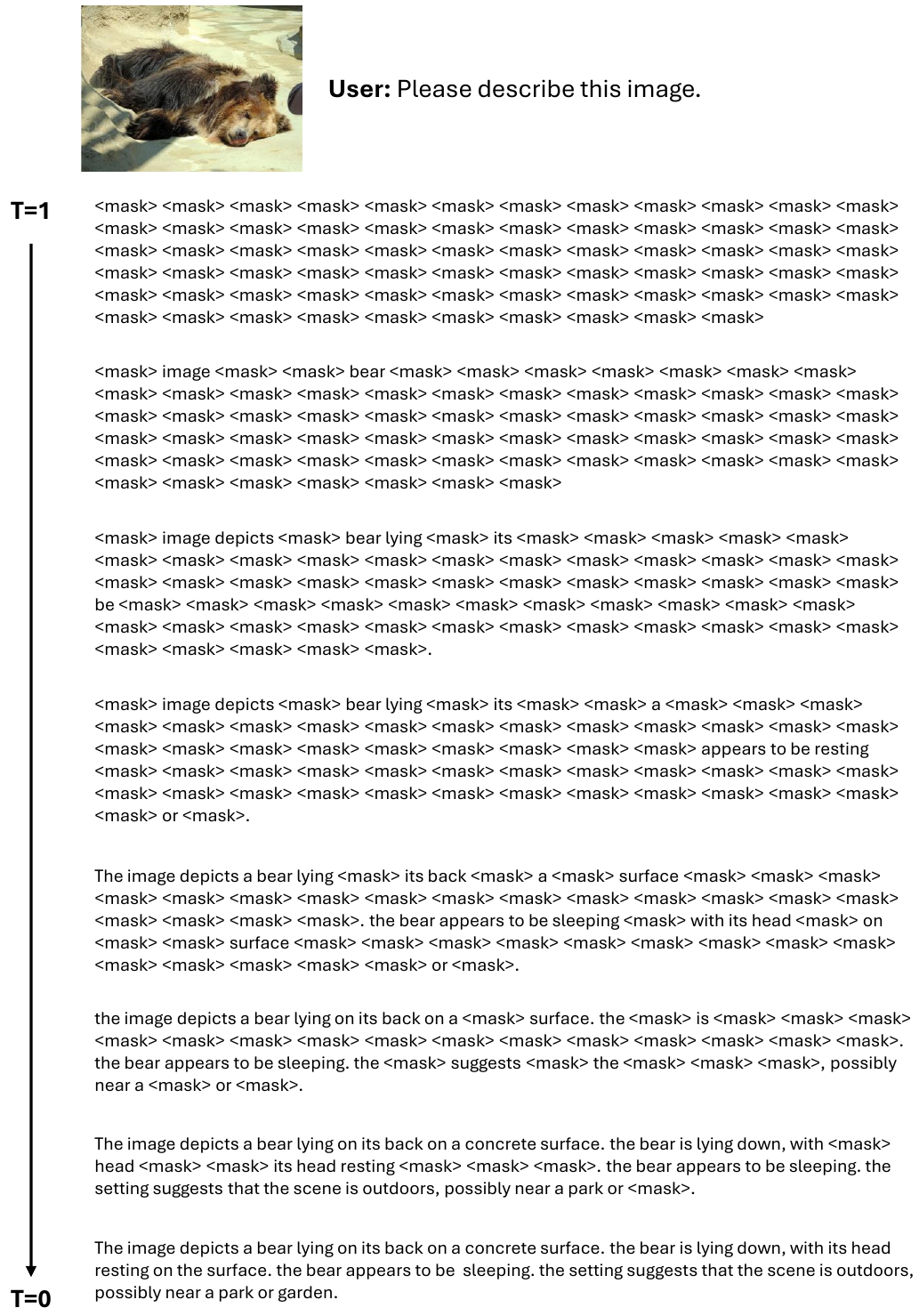}
    \caption{Image-to-text generated results in each step.}
    \label{fig:bear_sbs}
\end{figure}

\section{Additional Experimental Results}
\label{sec:additional_exp_res}

We provide a detailed breakdown of the MME benchmark results in Tab.~\ref{tab:mme_detail}. Muddit demonstrates strong performance in existence, color, and scene understanding, while also exhibiting solid reasoning capabilities.

\begin{table}[h]
\centering
\caption{Detailed MME results.}
\label{tab:mme_detail}
\renewcommand{\arraystretch}{1.15}
\setlength{\tabcolsep}{6pt}
\small
\begin{tabular}{l l r}
\toprule
\textbf{Category} & \textbf{Task} & \textbf{Score} \\
\midrule

\multirow{11}{*}{\textbf{Perception}}
  & Existence & 135.00 \\
  & Count & 78.33 \\
  & Position & 53.33 \\
  & Color & 140.00 \\
  & Posters & 62.24 \\
  & Celebrity & 56.18 \\
  & Scene & 107.25 \\
  & Landmark & 94.50 \\
  & Artwork & 76.00 \\
  & OCR & 52.50 \\
  \addlinespace[2pt]
  & \textbf{Total} & \textbf{855.34} \\
\midrule

\multirow{5}{*}{\textbf{Cognition}}
  & Commonsense Reasoning & 78.57 \\
  & Numerical Calculation & 90.00 \\
  & Text Translation & 57.89 \\
  & Code Reasoning & 57.50 \\
  \addlinespace[2pt]
  & \textbf{Total} & \textbf{283.97} \\
\bottomrule
\end{tabular}
\end{table}

\section{Additional Ablation Studies}
\label{sec:additional_ablation_study}

\subsection{Ablation study on the CFG for image-to-text generation}
As shown in Tab.~\ref{tab:cfg_ablation}, we report performance on MS-COCO captioning and VQAv2 benchmarks. Moderate CFG values (e.g., 1.5) yield the best results, while higher scales lead to degraded performance.

\begin{table}[t]
\centering
\caption{Ablation study on the effect of classifier-free guidance (CFG) scale.}
\begin{tabular}{lccccc}
\toprule
Dataset & CFG = 1 & CFG = 1.5 & CFG = 2 & CFG = 2.5 & CFG = 3 \\
\midrule
MS-COCO & 57.2 & 59.9 & 58.2 & 51.3 & 47.2 \\
VQAv2   & 65.8 & 68.2 & 64.7 & 55.4 & 49.2 \\
\bottomrule
\end{tabular}
\label{tab:cfg_ablation}
\end{table}

\begin{table}[t]
\centering
\caption{Comparison of model efficiency across different resolutions and steps. We report throughput for both text-to-image generation (images per second) and image-to-text tasks (tokens per second). Muddit achieves the best overall balance, matching the highest text-to-image throughput while significantly outperforming others in image-to-text speed.}
\label{tab:model_efficiency}
\begin{tabular}{lcccc}
\toprule
Model & Image Res & Steps & Text-to-Image (img/s) & Image-to-Text (token/s) \\
\midrule
Meissonic & 1024 & 32 & 0.23 & -- \\
UniDisc   & 512  & 32 & 0.89 & 79.36 \\
Monetico  & 512  & 32 & 1.00 & -- \\
D-DiT     & 512  & 28 & 0.62 & 26.89 \\
Muddit    & 512  & 32 & 1.00 & 99.98 \\
\bottomrule
\end{tabular}
\end{table}

\section{Inference Time Analysis}
\label{sec:inference}

\begin{figure}[t]
    \centering
    \includegraphics[width=.8\linewidth]{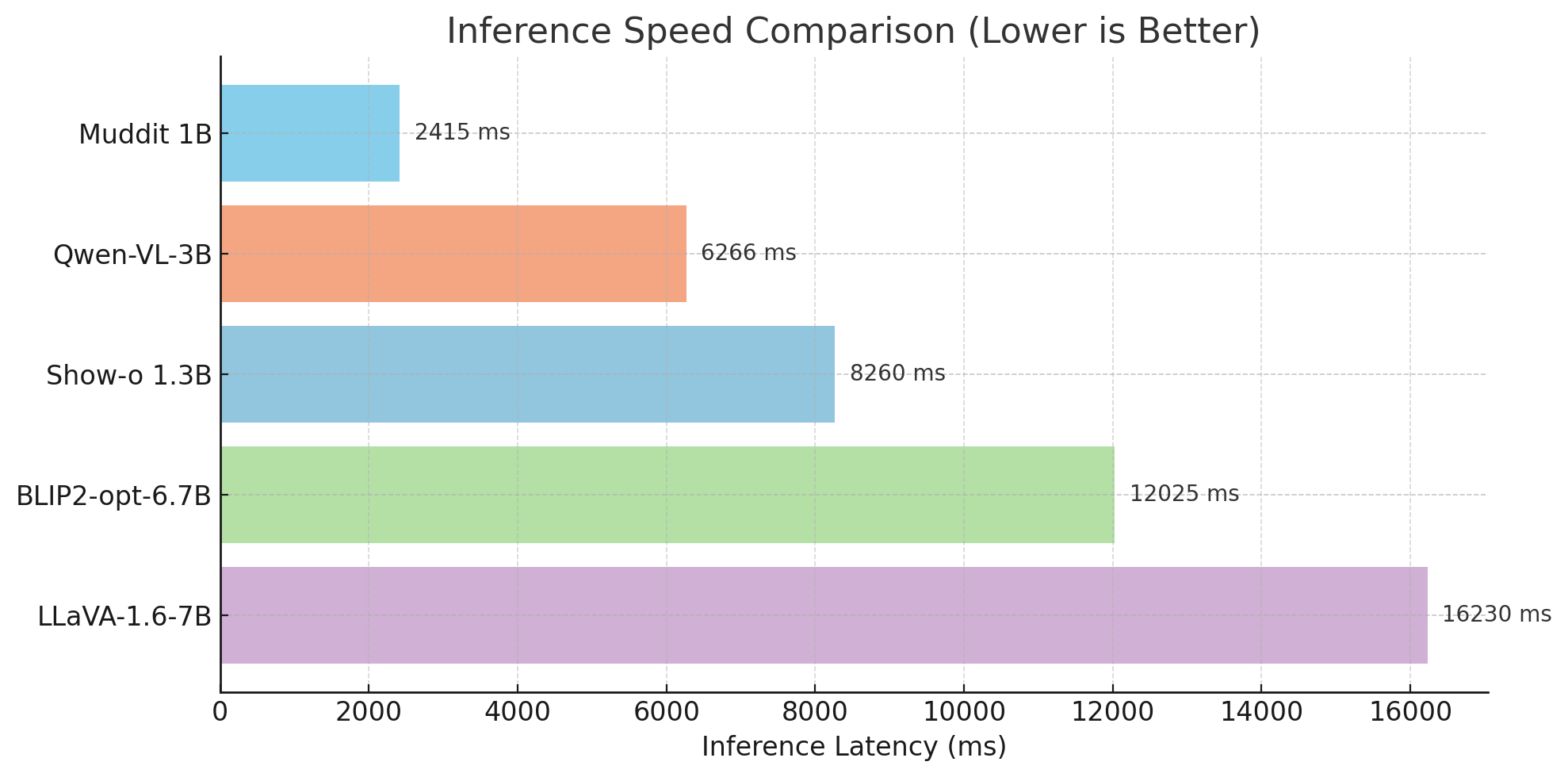}
    \caption{Inference speed comparison. We use 32 inference steps for \name{} and fix the sequence length to 77 across all models.}
    \label{fig:speed}
\end{figure}

As shown in Fig.~\ref{fig:speed}, autoregressive multimodal models are inherently limited by token-by-token decoding, which constrains their inference speed. \name{} overcomes this bottleneck with a parallel discrete diffusion decoder, reducing average latency to just 1.49 seconds, achieving a 4$\times$ to 11$\times$ speedup over competitive baselines (4.2$\times$ faster than Qwen-2.5-VL, 5.6$\times$ than Show-o, 8.1$\times$ than BLIP-2, and 10.9$\times$ than LLaVA-1.6).

We also present detailed FLOPs comparison between Autoregressive and Discrete Diffusion.

\noindent
\textbf{Autoregressive (AR) without KV Cache:}
\begin{itemize}
  \item At step $t$, the model attends over $t$ previous tokens.
  \item Per-step attention FLOPs: $O(t^2 D)$.
  \item Total FLOPs:
  \[
  \sum_{t=1}^{L} O(t^2 D) = O\left(D \sum_{t=1}^{L} t^2 \right) = O\left(D \cdot \frac{L(L+1)(2L+1)}{6} \right) = O(L^3 D)
  \]
\end{itemize}

\noindent
\textbf{Autoregressive (AR) with KV Cache:}
\begin{itemize}
  \item At step $t$, Q is computed for 1 token, and attends to $t$ K/V keys.
  \item Per-step attention FLOPs: $O(t D)$.
  \item Total FLOPs:
  \[
  \sum_{t=1}^{L} O(t D) = O\left(D \sum_{t=1}^{L} t \right) = O\left(D \cdot \frac{L(L+1)}{2} \right) = O(L^2 D)
  \]
\end{itemize}

\noindent
\textbf{Discrete Diffusion:}
\begin{itemize}
  \item Each step updates the full sequence (length $L$) in parallel.
  \item Per-step attention FLOPs: $O(L^2 D)$.
  \item Total FLOPs:
  \[
  T \cdot O(L^2 D) = O(T L^2 D), \quad T \ll L
  \]
\end{itemize}

\noindent
While discrete diffusion may appear less efficient than autoregressive (AR) models with KV caching in terms of theoretical FLOPs, it offers a significant advantage over AR without caching—achieving an L/T speedup by updating the full token sequence in parallel over T iterations. In practice, the higher degree of parallelism leads to competitive, and often faster, inference speed compared to AR models, especially when considering real-world GPU throughput. As KV cache techniques for discrete diffusion are rapidly evolving~\citep{ma2025dkv}, we expect further acceleration in the near future, narrowing the theoretical speed gap even with KV-cache AR baselines.

In Tab.~\ref{tab:model_efficiency}, we compared Muddit against other non-autoregressive models, running all tests on a single A800 80 GB GPU. Muddit demonstrated a clear advantage in both image and text generation.

\section{Generated Results Step by Step}
\label{sec:step}
\name{} frames text generation as reverse discrete diffusion over a fixed-length sequence of 77 token indices. At inference time, the model performs $16 \leq T \leq 32$ denoising steps, starting from a maximally entropic prior where every token is masked. At each step $t$, a parameter-shared transformer $\mathtt{G}$ predicts a categorical distribution over all positions in parallel, and a sampler $\mathtt{S}$ selects the next sequence:
\begin{equation}
\mathbf{x}_{t-1} = \mathtt{S}(\mathtt{G}(\mathbf{x}_{t}, \mathtt{c}, t), \mathbf{x}_t, t),
\qquad t = T,\ldots,1,
\end{equation}
where $\mathbf{x}_t \in \mathbb{V}^{77}$ is the token sequence at step $t$, and $\mathtt{c}$ denotes conditioning inputs. The logits can be tempered or top-$k$ filtered before sampling each token independently. The resulting sequence $\mathbf{x}_{t-1}$ seeds the next step, enabling fast, parallel decoding without autoregressive constraints.

Because all positions are updated in parallel, \name{} preserves global syntactic and semantic structure throughout the reverse diffusion process—unlike left-to-right autoregressive models, which can only condition on past predictions. Empirically, as few as $16 \leq T \leq 32$ steps are sufficient to approximate the natural language distribution with high fidelity. Thus, \name{} unifies diffusion generation with parallel decoding, effectively overcoming the serial bottleneck that limits conventional autoregressive multimodal models.

We present two examples in Fig.~\ref{fig:room_sbs} and \ref{fig:bear_sbs}.

\section{Discussion}
\label{sec:discussion}
\subsection{Limitations}
While \name{} advances discrete diffusion for unified multimodal generation, it still has several limitations. First, due to its token-level discrete representation, the model may underperform continuous diffusion models in generating photorealistic or high-resolution images. Second, \name{} is initialized from a pretrained text-to-image foundation model, which offers strong visual priors but limits its capacity for rich text understanding and generation compared to the latest large language models. This makes it less suitable for tasks that require long-form understanding and generation or deep linguistic reasoning. 

\subsection{Broader Impacts}
\name{} explores a new paradigm in multimodal generation by leveraging a strong visual prior as the backbone, in contrast to the prevailing trend of scaling large language models. This offers a complementary path toward efficient, grounded multimodal generation, particularly in vision-centric applications. The model’s ability to generate aligned visual and textual outputs in a fast, parallel manner could benefit downstream tasks, especially in completion-based scenarios such as masked captioning, image editing, and code implementation. However, as with all generative models, there remains a risk of misuse in synthetic content creation.

\section{The Use of Large Language Models}
\label{sec:usage_llm}
During the preparation of this paper, large language models were used only for language polishing and minor editing. 
All research ideas, methods, and experimental results were carried out entirely by the human authors.

\end{document}